\def\year{2021}\relax
\documentclass[letterpaper]{article} % DO NOT CHANGE THIS
\usepackage{aaai21} % DO NOT CHANGE THIS
\usepackage{times} % DO NOT CHANGE THIS
\usepackage{helvet} % DO NOT CHANGE THIS
\usepackage{courier} % DO NOT CHANGE THIS
\usepackage[hyphens]{url} % DO NOT CHANGE THIS
\usepackage{graphicx} % DO NOT CHANGE THIS
\usepackage{natbib} % DO NOT CHANGE THIS AND DO NOT ADD ANY OPTIONS TO IT
\usepackage{caption} % DO NOT CHANGE THIS AND DO NOT ADD ANY OPTIONS TO IT
\nocopyright
\usepackage{amsmath}
\usepackage{amssymb}
\usepackage{amsthm}
\usepackage{enumerate}
\usepackage{booktabs}
\usepackage[switch]{lineno}
\usepackage{xcolor}
\usepackage{xspace}

\usepackage{ifthen}
\newboolean{proofs}
\setboolean{proofs}{true} % flip to include/exclude proofs
\newcommand{\Proof}[1]{\ifthenelse{\boolean{proofs}}{\begin{proof}\color{black} #1 \end{proof}}{}}
\newcommand{\RProof}[1]{\ifthenelse{\boolean{proofs}}{\begin{proof}[\textcolor{magenta}{Proof}]\color{magenta} #1 \end{proof}}{}}

\usepackage{tikz}
\usetikzlibrary{arrows.meta, calc, positioning, backgrounds}
\tikzset{%
  diagonal fill/.style 2 args={fill=#2, path picture={
    \fill[#1, sharp corners] (path picture bounding box.south west) -|
                             (path picture bounding box.north east) -- cycle;}},
  reversed diagonal fill/.style 2 args={fill=#2, path picture={
    \fill[#1, sharp corners] (path picture bounding box.north west) |-
                             (path picture bounding box.south east) -- cycle;}}
}

\newcommand{\Omit}[1]{}
\newcommand{\denselist}{\itemsep -1pt\partopsep 0pt}

\newcommand{\citeay}[1]{\citeauthor{#1} (\citeyear{#1})}

\newcommand{\LL}{\mathcal{L}}

\newtheorem{definition}{Definition}
\newtheorem{def-thm}[definition]{Definition and Theorem}
\newtheorem{thm-def}[definition]{Theorem and Definition}
\newtheorem{theorem}[definition]{Theorem}

\newtheorem*{definition*}{Definition}
\newtheorem*{def-thm*}{Definition and Theorem}
\newtheorem*{thm-def*}{Theorem and Definition}
\newtheorem*{theorem*}{Theorem}
\newtheorem*{lemma*}{Lemma}
\newtheorem*{proposition*}{Proposition}
\newtheorem*{corollary*}{Corollary}

\def\eob{{ % end-of-block symbol: complex stuff, taken from QED.sty, do not modify except for symbol below
  \parfillskip=0pt % so \par doesnt push \square to left
  \widowpenalty=10000 % so we don't break the page before \square
  \displaywidowpenalty=10000 % ditto
  \finalhyphendemerits=0 % TeXbook exercise 14.32
  \leavevmode % \nobreak means lines not pages
  \unskip % remove previous space or glue
  \nobreak % don't break lines
  \hfil % ragged right if we spill over
  \penalty50 % discouragement to do so
  \hskip.2em % ensure some space
  \null % anchor following \hfill
  \hfill % push \square to right
  \raisebox{0pt}{\tikz{\draw[fill=black] (0,0) rectangle +(.5ex,1.5ex);}} % actual symbol
  \par}} % build paragraph

\newenvironment{example}{\noindent\textbf{Example.}\,}{\eob}
\newenvironment{example-no-eob}{\noindent\textbf{Example.}\,}{}

\newenvironment{linearassumption}{\noindent\textbf{Linear features assumption.}\,}{\eob}

\newcommand{\CHECK}[1]{\textcolor{red}{*** CHECK: #1 ***}}

\newcommand{\Q}{\mathcal{Q}}

\newcommand{\pplus}{\hspace{-.05em}\raisebox{.15ex}{\footnotesize$\uparrow$}}
\newcommand{\mminus}{\hspace{-.05em}\raisebox{.15ex}{\footnotesize$\downarrow$}}

\newcommand{\EQ}[1]{#1{\,=\,}0}
\newcommand{\GT}[1]{#1{\,>\,}0}

\newcommand{\DEC}[1]{#1\mminus}
\newcommand{\INC}[1]{#1\pplus}
\newcommand{\UNK}[1]{#1?}

\newcommand{\prule}[2]{\{ #1 \} \mapsto \{ #2 \}}

\newcommand{\iw}[1]{\ensuremath{\text{IW}(#1)}\xspace}
\newcommand{\iwf}[1]{\ensuremath{\text{IW}_{#1}}\xspace}
\newcommand{\siw}{\ensuremath{\text{SIW}_{\Phi}}\xspace}
\newcommand{\siwR}{\ensuremath{\text{SIW}_{\text R}}\xspace}
\title{General Policies, Serializations, and Planning Width$^*$}

\author{Blai Bonet,\textsuperscript{\rm 1} Hector Geffner\textsuperscript{\rm 2} \\}
\affiliations {
  \textsuperscript{\rm 1}\,Universitat Pompeu Fabra, Barcelona, Spain \\
  \textsuperscript{\rm 2}\,ICREA \& Universitat Pompeu Fabra, Barcelona, Spain \\
  bonetblai@gmail.com, hector.geffner@upf.edu
}

\makeatletter
\def\blfootnote{\gdef\@thefnmark{}\@footnotetext}
\makeatother

\begin{document}
%\linenumbers
\allowdisplaybreaks

\maketitle

\begin{abstract}
  It has been observed that in many of the benchmark planning domains, atomic goals can be reached
  with a simple polynomial exploration procedure, called IW, that runs in time exponential in the problem width.
  Such problems have indeed a {bounded width}: a width that does not grow with the number of problem
  variables and is often no greater than two. Yet, while the notion of width has become part of
  the state-of-the-art planning algorithms like BFWS, there is still no good explanation for why so many
  benchmark domains have bounded width. In this work, we address this question by
  relating bounded width and serialized width to ideas of generalized planning, where
  general policies aim to solve multiple instances of a planning problem all at once.
  We show that bounded width is a property of planning domains that admit
  optimal general policies in terms of features that are explicitly or implicitly
  represented in the domain encoding. The results are extended to the larger class of domains
  with bounded serialized width where the general policies do not have to be optimal.
  The study leads also to a new simple, meaningful, and expressive language for specifying
  domain serializations in the form of \textbf{policy sketches}
  which can be used for encoding domain control knowledge by hand or for learning it from traces.
  The use of sketches and the meaning of the theoretical results are all illustrated through a
  number of examples.\blfootnote{$^*$This paper is the long version of \cite{bonet:width}.}
\end{abstract}

\section{Introduction}

Pure width-based search methods exploit the structure of states to enumerate the state space in ways
that are different from standard methods like breadth-first, depth-first, or random search \cite{nir:ecai2012}.
For this, width-based methods appeal to a notion of novelty to establish a preference for first visiting states that
are most novel. Novelty-based methods have also been used in the context of genetic algorithms
where a greedy focus on the function to optimize (fitness) often leads to bad local optima \cite{lehman:novelty,lehman:novelty2},
and in reinforcement learning to guide exploration in large spaces where reward is
sparse \cite{rl-exploration1,rl-exploration2,rl-exploration3}.

In classical planning, i.e., planning in factored spaces for achieving a given goal from a known initial state
\cite{geffner:book,ghallab:book}, the notion of novelty is now part of state-of-the-art search algorithms
like BFWS \cite{nir:icaps2017,nir:aaai2017} and has been applied successfully in purely exploratory settings
where no compact model of the actions or goals is assumed to be known a priori \cite{guillem:ijcai2017,bandres:aaai2018}.
The basic width-based planning algorithms are simple and they all assume that the states are factored, assigning
values to a fixed number of boolean features $F$ that in classical planning is given by the atoms in the problem.
The procedure \iw{1} is indeed a breadth-first search that starts in the given initial state and prunes
all the states that do not make a feature from $F$ true for the first time in the search.
\iw{k} is like \iw{1} but using the set of features $F^k$ that stand for conjunctions of up to $k$ features from $F$.
For many benchmark domains, it has been shown that \iw{k} for a small and constant value of $k$ like $k=2$,
called the domain width, suffices to compute plans, and indeed optimal plans, for any atomic goal \cite{nir:ecai2012}.
Other algorithms like BFWS make use of this property for serializing conjunctive goals into atomic ones; an idea
that is also present in algorithms that precompute atomic landmarks \cite{hoffmann:landmarks},
and in particular, those that use landmark counting heuristics \cite{lama}.

A key open question in the area is why these width-based methods are effective at all, and in particular, why
so many domains have a small width when atomic goals are considered. Is this a property of the domains?
Is it an accident of the manual representations used? In this work, we address these and related questions.
For this, we bring the notion of \textbf{general policies}; policies that solve multiple instances of a planning
domain all at once \cite{srivastava08learning,bonet:icaps2009,hu:generalized,BelleL16,anders:generalized},
in some cases by appealing to a fixed set $\Phi$ of general state features that we restrict to be {linear}.
% i.e., the value of a feature $\phi(s)$ in a state must be computed in time that is
% linear in the number of atoms $N$ in the problem and the number of such values must be linear in $N$
% as well. Some of the formal results to be developed
A number of correspondences are then obtained connecting the notions of \emph{width,
the size of general policies as measured by the number of features used,
and the more general and useful notion of serialized width;} i.e.,
width under a given serialization. The study leads us to formalize
the abstract notion of serialization and to analyze the conditions
under which serializations are well-formed. Moreover, from the relations
established between general policies and serializations, we obtain
a new simple, meaningful, and expressive language for specifying
serializations, called \textbf{policy sketches}, which can be used for encoding
domain control knowledge by hand or for learning it from traces.
While exploring these potential uses of sketches is beyond the scope
of this paper, the use of sketches and the meaning of the theoretical
results are all illustrated through a number of examples.

The paper is organized as follows. We review first the notions of width, representations, and general policies,
and relate width with the size of such policies, as measured by the number of features. Then we
introduce serializations, the more general notion of serialized width, the relation between general policies
and serialized width, and finally, policy sketches and their properties.

\section{Width}

\iw{1} is a simple search procedure that operates on a rooted directed graph where the nodes represent
states, and states assign values $v$ to a given set $F$ of features $f$ \cite{nir:ecai2012}.
\iw{1} performs a breadth-first search starting at the root but pruning the states that do not make
an atom $f{=}v$ true for the first time in the search. For classical planning
problems expressed in languages such as (grounded) STRIPS, the features $f$ are the problem variables
which can take the values true or false. In other settings, like the Atari games as supported in ALE \cite{ale},
the features and their values are defined in other ways \cite{nir:ijcai2015,bandres:aaai2018}.
The procedure \iw{k} for $k > 1$ is \iw{1} but with a feature set given by $F^k$. % ($k$-tuples over $F$).

A finding reported by \citeay{nir:ecai2012} and exploited since in width-based algorithms is
that the procedure \iw{k} with $k=2$ suffices to solve a wide variety of planning problems.
Indeed, Lipovetzky and Geffner consider the 37,921 instances that result from all the domains
used in planning competitions until 2012, where each instance with a goal made up of a conjunction of $k$
atoms is split into $k$ instances, each one with a single atomic goal. They report that \iw{2}
solves more than 88\% of such instances, and moreover, 100\% of the instances from 26 of the 37
domains considered.

This is remarkable because \iw{k} when $k$ is smaller
than the number $n$ of problem variables is an \emph{incomplete procedure.} Indeed, while breadth-first
search expands a number of nodes that is exponential in $n$, \iw{2} expands
a quadratic number of nodes at most. Moreover, the results are not an
accident resulting from a lucky choice of the instances, as one can formally prove
that \iw{2} solves \textbf{any} instance of many of these domains when
the goal is a single atom.

\Omit{
  For example, \iw{2} solves any Blocksworld
  instance where the goal is of the form $on(x,y)$.
  These instances thus define exponential state spaces
  that have (optimal) solutions in a small polynomial envelope
  that can be found in low polynomial time. Indeed, since the pruning (novelty) test
  in \iw{k} runs in time that is exponential in $k-1$ when the
  actions effects are known and bounded in number (as when the actions are represented
  in STRIPS), \iw{k} runs in $exp(2k-1)$ time, so that \iw{2} runs in cubic time.
} % omit

Underlying the IW algorithms is the notion of \emph{problem width}, which borrows from
similar notions developed for constraint satisfaction problems
and Bayesian networks, that are intractable but have algorithms
that run in time and space that are exponential in the treewidth of the
underlying graphs \cite{freuder:width,pearl:book,rina:width}.
For planning, the notion introduced by Lipovetzky and Geffner
takes this form:\footnote{For convenience, we set the width of problems that can be solved
  in one step to zero.
}
% The definition is the one from Lipovetzky and Geffner,
% except that for convenience, the width of a problem is set to zero
% when it can be solved in one step:

\begin{definition}[Based on \citeauthor{nir:ecai2012}, 2012]
  \label{def:width}
  The \textbf{width} $w(P)$ of problem $P$ is the minimum $k$ for which there is a
  \textbf{sequence $t_0,t_1,\ldots,t_m$ of atom tuples $t_i$},
  each with at most $k$ atoms, such that:
  \begin{enumerate}[1.]
    \item $t_0$ is true in the initial state of $P$,
    \item any optimal plan for $t_i$ can be extended into an optimal plan for $t_{i+1}$ by adding a single action, $i=1,\ldots,n-1$,
    \item any optimal plan for $t_m$ is an optimal plan for $P$. % , or can be extended into such by adding a single action.
  \end{enumerate}
  The width is $w(P)=0$ iff the initial state of $P$ is a goal state.
  For convenience, we set $w(P)$ to $0$ if the goal of $P$ is reachable in a single step,
  and to $w(P)=N+1$ if $P$ has no solution where $N$ is the number of atoms in $P$.
\end{definition}

Chains of tuples $\theta=(t_0,t_1, \ldots, t_m)$ that comply with conditions 1--3 are called \textbf{admissible},
and the size of the chain is the size $|t_i|$ of the largest tuple in the chain.
The width $w(P)$ is thus $k$ if $k$ is the minimum size of an admissible chain for $P$.
Notice that the definition of width does not presuppose a particular language for
specifying the actions or goals, which can actually be specified procedurally.
It just assumes that states are factored and assign truth values to a pool of atoms.
The tuples $t_i$ in an admissible chain can be regarded as subgoals or stepping stones
in the way to the goal of $P$ that ensure that it will be reached optimally.
The main properties of the \iw{k} algorithm can then be expressed as follows:\footnote{It
  is assumed that the number of atoms  affected by an action is bounded by a constant.
  When this is not the case, the time bound in Theorem~\ref{thm:width:nir} becomes $O(bN^{2k})$.
}

\begin{theorem}[\citeauthor{nir:ecai2012}, 2012]
  \label{thm:width:nir}
  \iw{k} expands up to $N^k$ nodes, generates up to $bN^k$ nodes, and runs in time and space $O(bN^{2k-1})$ and $O(bN^k)$, respectively,
  where $N$ is the number of atoms and $b$ bounds the branching factor in problem $P$.
  \iw{k} is guaranteed to solve $P$ optimally (shortest path) if $w(P) \leq k$.
  %IW($k$) expands up to $n^k$ nodes, generates up $b n^k$ nodes, and runs in time $exp(2k-1)$ in factored encodings
  %where $n$ is the number of problems variables, and $b$ is the branching factor (average number of children per node).
  %It's guaranteed to solve $P$ optimally (shortest path) if $w(P) \leq k$.
\end{theorem}
\Proof{
  Since the number of tuples of size at most $k$ is bounded by $N^k$,
  \iw{k} expands up to $N^k$ nodes and generates up to $bN^k$ nodes,
  where each expansion takes time bounded by $b$.
  Under the assumption that each transition flips the value of up to $M$ (constant)
  number of atoms, each dequed state can make true up to ${N \choose k} - {N-M \choose k}=O(N^{k-1})$
  new tuples of atoms of size at most $k$.
  Seen tuples are stored in a \emph{perfect hash} of size $O(N^k)$ which supports
  operations in constant time.
  Therefore, the test for expansion for each dequed state takes time $O(N^{k-1})$,
  and thus the total time incurred by \iw{k} for exploring the search tree is $O(bN^{2k-1})$.
  From the definition of width, it is not hard to see that if $w(P)\leq k$, then \iw{k}
  generates a node for a goal state, and the path from the initial state to the
  first goal state that is generated is an optimal (shortest) path.
}

When the width of problem $P$ is not known, the \textbf{IW algorithm}
can be run instead which calls \iw{k} iteratively for
$k\!=\!0,1,\ldots,N$ until the problem is solved, or shown to have no solution
when \iw{N} finds no plan.
While these algorithms are not aimed at being practical as planning algorithms,
some state-of-the-art planners, make use of these ideas in a slightly different
form \cite{nir:icaps2017,nir:aaai2017}.

\Omit{
  It is useful to revise how one proves that an infinite collection of problems has bounded width.
} % omit

\smallskip

\begin{example}
  Let $\Q_{clear}$ be the class of Blocksworld problems $P$ in the standard stack/unstack encoding
  whose goal is to achieve the atom $clear(x)$ and an empty gripper, starting with $clear(x)$ false
  and an empty gripper. Let $B_1, \ldots, B_m$ for $m > 0$ be the blocks above $x$ from top to bottom
  in $P$, and let us consider the {chain of subgoals} $t_0, \ldots, t_{2m-1}$ where
  $t_0 = \{clear(B_1)\}$, $t_{2i-1} = \{hold(B_i)\}$, and $t_{2i}=\{ontable(B_i)\}$,
  $i=1, \ldots, m$.
  It is easy to check that conditions 1--3 above hold for this chain, hence, $w(P) \leq 1$.
  Since $w(P) > 0$, as the goal cannot be reached in zero or one step in general, $w(P)=1$.
  %Since $w(P) \neq 0$ as the goal is not true in the initial state and
  %cannot be reached in one step, $w(P)=1$.
\end{example}

\smallskip

\begin{example}
  Let $\Q_{on}$ be the class of Blocksworld problems $P$ where the
  goal is $on(x,y)$. For simplicity, let us assume % and without loss of generality, let us assume
  that $x$ and $y$ are not initially in the same tower,
  and there are blocks $B_1, \ldots, B_m$ above $x$, and blocks $D_1, \ldots, D_\ell$ above $y$.
  It is not difficult to check that the chain $t_0,\ldots,t_{2m},t'_0,\ldots,t'_{2\ell},t''_0,t''_1$
  is admissible and of size 2,
  where $t_i$ for $i=0,\ldots,2m$ is as in previous example,
  $t'_{2i}=\{hold(D_i),clear(x)\}$ and $t'_{2i-1}$ $=\{ontable(D_i),$ $clear(y)\}$ for
  $i=0,\ldots,\ell$, $t''_0=\{hold(x),clear(y)\}$, and $t''_1=\{on(x,y)\}$.
  %is an admissible chain of size 2.
  Then, $w(P)=2$ since $w(P)\not\leq 1$.
  %%For proving $w(P) \leq 2$, we consider the chain
  %\[t_0, \ldots, t_{2m}, \ldots, t_{2(m+l)}, t_{2(m+l)+1}, t_{2(m+l+1)}\]
  %\noindent where the $t_i$ subgoals are:
  %%
  %\[\{clear(B_1)\} \ , \ldots, \ \{hold(B_i)\} \ , \ \{ontable(B_i)\} \ , \ldots,
  %\{hold(D_i),clear(x)\} \ , \ \{ontable(D_i),clear(x),\} \ , \ldots, \]
  %\noindent ending with $\{hold(x),clear(y)\}$ and $\{on(x,y)\}$, where
  %$i=1,\ldots,m$ and $j=1, \ldots, l$.
  %
  %One can check that this chain of subgoals is \textbf{admissible} (complies with conditions 1--3 above), and hence $w(P) \leq 2$,
  %In addition since $w(P) \not\leq 1$, it follows that $w(P)=2$.
  %The intuition for why $w(P)\not=1$ is that there is no chain of atomic subgoals goals $t_i$
  %that can take us step by step to the goal $on(x,y)$ of $P$ while complying with conditions 1--3. Indeed, the
  %atoms $clear(x)$ and $clear(B)$ must be achieved \textbf{jointly} in the way to the goal $on(x,y)$,
  %but that conjunction does not have width $1$ either. Indeed, IW(1) can achieve the subgoal $clear(x) \land clear(B)$ only
  %by ``luck'', not by structural reasons, as when it achieves $clear(x)$ first and then puts away the blocks above $B$
  %somewhere else. By saying that the width of $P$ is $2$, IW(2) will solve $P$ no matter how the children of a state
  %are ordered during the breadth-first search underlying IW.
\end{example}

\Omit{
  While the notion of width and the associated notion of novelty have become part of
  state-of-the-art planning algorithms, there is however no explanation
  as to \textbf{why} the notion is useful at all, and \textbf{why} so many standard planning benchmarks
  turn out to have small bounded width (for atomic goals). Is this an accident? Is it related to the particular choice
  of the benchmarks? This is the question that we address in the rest of the paper.

  The answer for short is the following: many of these benchmarks domains are simple in
  the sense that there is a general strategy that can be used to solve them
  that can be executed in polynomial time and can expressed in a compact
  form through the use of a fixed number of general features.
  Under such conditions, there are compact planning representation of the domain
  instances that have bounded widths: widths bounded indeed by the
  number of features used in the general policies.
} % omit

\section{Representations}

The width of a planning problem is tied to the representation language
and the encoding of the problem in the language.
For example, the problem of moving a number of packages $N$ from one room to the next,
one by one, has a width that grows with $N$ in standard encodings where each
package has a name, but width $2$ when the packages are indistinguishable
from each other and the number of packages is encoded in unary
(with one atom per counter value).\footnote{The problem would still not have
  bounded width if the counter is represented in binary using a logarithmic number
  of atoms.
}

In order to deal with a variety of possible languages and encodings, and since
width-based methods rely on the \emph{structure of states} but not on the \emph{structure of actions} (i.e.,
action preconditions and effects), we consider \emph{first-order languages} for describing states
in terms of atoms that represent objects and relations, leaving out from the language the representation of action
preconditions and effects. It is assumed that {the possible state transitions} $(s,s')$
from a state $s$ are a \emph{function of the state} but no particular representation of this
function is assumed. In addition, the state language is extended with \emph{features} $f$ whose
values $f(s)$ in a state $s$ are determined by the state. The features provide additional expressive power
and ways for bridging different state representation languages, although they are logically redundant
as their value is determined by the truth value of the atoms in the state. The features extend the notion of
\emph{derived predicates} as defined in PDDL, as they do not have to be boolean,
and they do not have to be defined in the language of first-order logic or logic programs
\cite{axioms:pddl}, but can be defined via procedures. Domains, instances, and states
are defined as follows:

\begin{definition}[Domains, problems, and states]
  \label{def:domains-problems-states}
  A \textbf{domain} is a pair $D=(R,F)$ where
  $R$ is a set of \textbf{primitive predicate symbols} with their corresponding arities, and
  $F$ is a set of \textbf{features} defined in terms of the primitive predicates with their corresponding
  range of feature values.
  A \textbf{problem} $P$ over domain $D=(R,F)$ is a tuple $P=(D,O,I,G)$ where $O$ is a set of unique object
  names $c$ (objects), and $I$ and $G$ are sets of ground atoms that denote the \textbf{initial} and \textbf{goal} states of $P$.
  A ground atom $r(c_1, \ldots, c_{a(r)})$ is made of a predicate $r \in R$ and an object tuple in $O^{a(r)}$ for the
  arity $a(r)$ of $r$.
  A \textbf{state} $s$ over problem $P=(D,O,I,G)$ is a collection of ground atoms.
  The state $s$ is a \textbf{goal} if $G \subseteq s$, and a \textbf{dead end} if a goal state cannot be reached from $s$ in $P$.
  A state $s$ denotes a unique valuation for the ground atoms in $P$: $s\vDash r(c_1,\ldots,c_k)$
  iff $r(c_1,\ldots,c_k)$ belongs to $s$.
\end{definition}

This is all standard except for the two details mentioned before: there are no action schemas,
and there are state features. For the former, it is implicitly assumed that in each problem
$P$, there is a function that maps states $s$ into the set of possible transitions $(s,s')$.
This implies, for example, that the states may contain \emph{static atoms,} like adjacency
relations, whose truth value are not affected by any action.
For the features, we make the assumption that they are \textbf{linear},
in the sense that they can be computed efficiently and only span a linear
number of values. More precisely:

\smallskip

\begin{linearassumption}
  The features $f$ in $F$ are either boolean or numerical, ranging in the latter case
  over the non-negative integers. The value of the feature $f$ in a state $s$ for problem $P$,
  $f(s)$, can be computed in time bounded by $O(bN)$ where $N$ is the number of atoms and
  $b$ bounds the branching factor in $P$. Numerical features can take up to $N$ values.
  %We assume a computational model where the arithmetic operations and comparisons over integers take constant time.
\end{linearassumption}

\smallskip

This assumption rules out features like $V^*(s)$ that stands for the optimal cost (distance) from $s$ to a goal
which may take a number of values that is not linear in the number of problem atoms, and whose computation may take
exponential time. In many cases, the features can be defined in the language of first-order logic
but this is not a requirement.

\Omit{
  For instance, in the Blocks world, $clear(x)$ can be defined as a boolean feature determined
  by the location of the blocks; i.e., $clear(x)$ iff $\neg \exists y. \, on(y,x)$, while counters $n(x)$
  can be defined as $|\phi(x)|$ for a formula $\phi(x)$ with $x$ as the only free variable,
  that counts the number of objects $c$ such that $\phi(c)$ is true in the state.
} % omit

\smallskip

\begin{example-no-eob}
  Three {state languages} for Blocksworld are:
  \begin{enumerate}[1.] \denselist
    \item $\LL_{BW}^1$ with the binary predicate (symbol) $on^2$ and the unary $ontable^1$ (superindex indicates arity),
    \item $\LL_{BW}^2$ with predicates $on^2$, $ontable^1$, $hold^1$, and $clear^1$,
    \item $\LL_{BW}^3$ with predicates $on^2$ and $hold^1$, and boolean features $ontable^1$ and $clear^1$. \eob
  \end{enumerate}
\end{example-no-eob}

\begin{example-no-eob}
  Four languages for a domain {Boxes}, where boxes $b$ containing marbles
  $ma$ must be removed from a table, and for this, the marbles in the box
  must be removed one by one first:
  \begin{enumerate}[1.] \denselist
    \item $\LL_{B}^1$ with predicates $ontable^1(b)$ and $in^2(ma,b)$,
    \item $\LL_{B}^2$ with predicates $ontable^1$, $in^2$, and $empty^1(b)$,
    \item $\LL_{B}^3$ with predicates $ontable^1$ and $in^2$, and features $n(b)$ that count the number of marbles in $b$,
    \item $\LL_{B}^4$ with predicates $ontable^1$ and $in^2$, and features $m$ and $n$ counting the number of marbles in a
      box with the least number of marbles, and the number of boxes left. \eob
  \end{enumerate}
\end{example-no-eob}

By abstracting away the details of the domain dynamics and the ability
to introduce features, it is simple to move from one state
representation to another.

The notion of width and the IW algorithms generalize to state
languages containing features in a direct fashion.
In both cases, the set of atoms considered is extended to contain
the possible feature values $f{=}v$ where $f$ is a
feature and $v$ is one of its possible values.
Features are logically redundant but
can have drastic effect on the problem width.

The width for class $\Q$ of problems $P$ over some domain $D$
is $k$, written as $w({\Q})=k$, if $w(P)=k$ for some $P \in \Q$
and $w(P') \leq k$ for every other problem $P'$ in $\Q$.

\smallskip

\begin{example}
  The width for the class of problems $\Q_{clear}$ where block $x$
  has to be cleared has width $1$ in the state languages
  $\LL_{BW}^i$, $i=1,2,3$, while the class $\Q_{on}$ has
  width $2$ for the three languages.
  On the other hand, for the class $\Q_{B_1}$ of instances from
  Boxes with a single box, the width is not bounded as it grows
  with the number of marbles when encoded in the languages
  $\LL_{B}^1$ and $\LL_B^2$, but it is $1$ when encoded in the
  languages $\LL_B^3$ or $\LL_B^4$.
  Likewise, for the class $\Q_{B}$ of instances from Boxes with
  arbitrary number of boxes, the encoding in the language $\LL_B^3$
  has width that is not bounded as it grows with the number of boxes, but remains
  bounded and equal to $2$ in $\LL_B^4$. %\footnote{See the supplemental material for the proofs.}
\end{example}

% omit

\section{Generalized policies}

We want to show that bounded width is a property of domains that admit
a certain class of general policies. Different language for expressing
general policies have been developed, some of which can deal with
relational domains where different instances involve different (ground) actions.
Most closely to this work, general policies have been
defined in terms of qualitative numerical planning problems (QNPs)
\cite{sid:aaai2011,bonet:ijcai2018,bonet:qnps}.
We build on this idea but avoid the introduction of QNPs by defining
policies directly as mappings from \emph{boolean feature conditions}
into \emph{feature value changes}.

A \textbf{boolean feature condition} for a set of features $\Phi$ is a condition
of the form $p$ or $\neg p$ for a boolean feature $p$ in $\Phi$, or $n=0$ or
$n>0$ for a numerical feature $n$ in $\Phi$.
% A set of boolean conditions is complete if each feature in $\Phi$ is mentioned in the set. %% Needed? HG
Similarly, a \textbf{feature value change} for $\Phi$ is an expression of the form
$p$, $\neg p$, or $\UNK{p}$ for a boolean feature $p$ in $\Phi$, and $\DEC{n}$,
$\INC{n}$, or $\UNK{n}$ for a numerical feature $n$ in $\Phi$.
% \CHECK{Shouldn't we say if $\DEC{n}$ in $E$, then $\GT{n}$ in $C$?}
% HG: not strictly needed --- Added it now 
General policies are given by a set of rules $C \mapsto E$
where $C$ and $E$ stands for boolean feature conditions and feature changes respectively.

\begin{definition}[Policies]
  \label{def:gen-policy}
  A \textbf{general policy} $\pi_\Phi$ for a domain $D$ over a set of features $\Phi$
  % defined in terms of a set of boolean and numerical features $\Phi$,
  % is a mapping from \emph{boolean feature conditions} into \emph{feature value changes} that is encoded as a set of
  % condition-effect rules of the form $C \mapsto E$, where $C$ is a set of boolean
  is a set of rules of the form $C \mapsto E$, where $C$ is a set of boolean
  feature conditions and $E$ is a set of feature value changes.
  The condition $\GT{n}$ is assumed in rules with effects $\DEC{n}$ or $\UNK{n}$.
  %A \textbf{general policy} $\pi_\Phi$ for a domain $D$ given a set of boolean and numerical features $\Phi$
  %is a mapping from \textbf{boolean feature conditions} into \textbf{feature value changes}
  %encoded as a set of condition-effect pairs of the form $C_i \mapsto E_i$, $i=1, \ldots, m$, where:
  %\begin{itemize}
  %  \item The \textbf{boolean feature conditions} in $C_i$ can be of the form $p$ or $\neg p$ for boolean features $p \in \Phi$,
  %    and $n=0$ or $n > 0$ for numerical features $n \in \Phi$,
  %  \item The \textbf{feature value changes} can be of the form $p$, $\neg p$, or $p?$ for boolean features $p \in \Phi$, and $n\mminus$, $n\pplus$, or
  %    $\UNK{n}$ for numerical features $n \in \Phi$
  %\end{itemize}
\end{definition}

\Omit{%% Doesn't seem to be needed and gets on the way
  In this representation, $\pi_\Phi(s)$ for a state $s$ over an instance $P$
  of a domain $D$ represents a collection of effects (feature value changes);
  namely, $\pi_\Phi(s) = \{ E_i \}_i$ where the index $i$ ranges over the
  conditions $C_i$ of the rules $C_i \mapsto E_i$ that are satisfied by the
  state $s$ (such a rule is referred to as \textbf{applicable} at $s$). % in the boolean valuation $\bar{\phi(s)}$.
} % omit

The policy $\pi_\Phi$ prescribes the possible actions $a$ to be done in a state $s$ over a problem $P$ \emph{indirectly,} as the
set of state transitions $(s,s')$ that the actions in $P$ make possible and which are \emph{compatible} with the policy:

\begin{definition}
  \label{def:transition}
  A transition $(s,s')$ \textbf{satisfies} the effect $E$ when:
  \begin{enumerate}[1.]
    \item if $p$ (resp.\ $\neg p$) in $E$, $p(s')=1$ (resp.\ $p(s')=0$),
    \item if $\DEC{n}$ (resp.\ $\INC{n}$) in $E$, $n(s) > n(s')$ (resp.\ $n(s) < n(s'))$, %and
    \item if $p$ (resp.\ $n$) is not mentioned at all in $E$, $p(s)=p(s')$ (resp.\ $n(s)=n(s')$).
  \end{enumerate}
  The transition $(s,s')$ is \textbf{compatible} with policy $\pi_\Phi$
  (or is a $\pi_\Phi$-transition) if there is a policy rule $C \mapsto E$
  such that $s$ makes true $C$ and $(s,s')$ satisfies $E$.
  %some effect $E$ in $\pi_\Phi(s)$.
  %% , and $\pi_\Phi$ is \textbf{applicable} at $s$
  %% if $\pi_\Phi(s)\neq\empty$.
\end{definition}

% A policy $\pi_\Phi$ for a class of problems $\Q$ is thus a restriction on the transitions $(s,s')$ that can be taken
% in the state $s$ of any instance $P$ in $\Q$.
Policy rules provide a description of how the value of the features must change along
the state trajectories that are compatible with the policy. Every transition $(s,s')$
in such trajectories have to be compatible with a policy rule $C \mapsto E$. 
The expressions $\UNK{p}$ and $\UNK{n}$ in $E$ stand for uncertain effects, meaning that $p$ and $n$ may change
in any way or not change at all. Features not mentioned in $E$, on the other hand, must keep
their values unchanged in a transition compatible with the rule $C \mapsto E$.

The definition does not exclude the presence of multiple policy rules $C \mapsto E$ with conditions $C$
that are all true in a state $s$. In such a case, for a state transition $(s,s')$ to be compatible with the policy,
the definition requires $(s,s')$ to satisfy one of the effect expressions $E$. On the other hand,
if the body $C$ of a single rule $C \mapsto E$ is true in $s$, for the transition to be compatible with the policy,
it must satisfy the effect $E$ of that rule.

\smallskip

\begin{example}
  A policy for solving the class $\Q_{clear}$
  can be expressed in terms of the features $\Phi=\{H,n\}$, where $H$
  is true if a block is being held, and $n$ counts the number of blocks above $x$.
  The policy can be expressed with two rules:
  \begin{equation}
    \label{pi:clear}
    \prule{\neg H, \GT{n}}{H, \DEC{n}} \ \ ; \ \ \prule{H, \GT{n}}{\neg H} \,.
  \end{equation}
  The first rule says that when the gripper is empty and there are blocks above $x$,
  an action that decreases $n$ and makes $H$ true must be chosen, while
  the second rule says % The second rule
  that when the gripper holds a block and there are blocks
  above $x$, an action that makes $H$ false and does not affect $n$ must
  be selected.
  % (notice that this rule is not compatible with actions that
  % place the block being held above $x$ as such will increase $n(x)$).
\end{example}

\smallskip

\Omit{
  A crucial advantage of this \emph{policy representation language} is
  that it is independent of both object names and the representation of
  the actions in $P$. The only requirement is that the features in $\Phi$
  can evaluated in any state (and that they are linear as defined above).
} % omit

The conditions under which a general policy $\pi_\Phi$ solves an instance $P$
and class $\Q$ are:

\begin{definition}[Trajectories and solutions]
  \label{def:trajectories-solutions}
  A \textbf{state trajectory} $s_0,\ldots,s_n$ for problem $P$ is \textbf{compatible}
  with policy $\pi_\Phi$ over features $\Phi$ (or is $\pi_\Phi$-trajectory),
  iff $s_0$ is the initial state of $P$, no state $s_i$ is goal, $0\leq i<n$, and each
  pair $(s_i,s_{i+1})$ is a possible state transition in $P$ that is compatible with $\pi_\Phi$.
  It is \textbf{maximal} if either $s_n$ is a goal state, no transition $(s_n,s_{n+1})$
  in $P$ is compatible with $\pi_\Phi$, or the trajectory is infinite (i.e., $n=\infty$).
  A policy $\pi_\Phi$ \textbf{solves} a problem $P$ if all {maximal state trajectories}
  $s_0, \ldots, s_n$ compatible with $\pi_\Phi$ are goal reaching (i.e., $s_n$ is a goal
  state in $P$). $\pi_\Phi$ solves a collection $\Q$ of problems if it solves each
  problem in $\Q$.
\end{definition}

It is easy to show that the policy captured by the rules in \eqref{pi:clear} solves $\Q_{clear}$.
The verification and {synthesis} of general policies of this type have been addressed
by \citeay{bonet:qnps} in the context of qualitative numerical planning, % problems} \cite{sid:aaai2011},
and by \citeay{bonet:aaai2019} where the set of features $\Phi$ and QNP model are learned
from traces.

\section{Generalized Policies and Width}

The first result establishes a relation between classes of problems that are
solvable by certain type of policies and their width. Namely, if there is a
``Markovian'' policy $\pi_\Phi$ that generates optimal plans for a class $\Q$,
the problems in $\Q$ can be encoded to have width bounded by $|\Phi|$.
% Let us make this formal and reveal all the assumptions behind this claim.
%%From now on we assume that the problems in $\Q$ are solvable
%%and that $\Q$ is ``closed'' in the sense that if $s$ is reachable
%%state from instance $P \in\Q$ that is not a dead end, then the
%%instance $P[s]$, i.e., $P$ but with initial state $s$, is also in $\Q$.
%%The assumption allows for dead-end states which can be avoided.

A policy $\pi_\Phi$ over the features $\Phi$ is Markovian in $\Q$ when
the features provide a suitable abstraction of the states; i.e., when
the possible next feature valuations are not just a function of the current
state, but of the feature valuation in the state.
% More precisely, if we say that a transition $(s,s')$ (resp.\ state $s$)
% in problem $P$ is \emph{reachable} by policy $\pi_\Phi$
% if there is a trajectory $\tau$ compatible with $\pi_\Phi$ that
% contains the transition (resp.\ state):
%%
% For formalizing this idea, let us say that a transition $(s,s')$ (resp.\ state $s$)
% in problem $P$ is reachable by policy $\pi_\Phi$ (or $\pi_\Phi$-reachable)
% where there is a trajectory $\tau$ compatible with $\pi_\Phi$ that
% contains the transition (resp.\ state).
More precisely, if we say that a state $s$ is \emph{optimal for a feature
valuation} $f$ in a problem $P$ when $f=f(s)$ and there is no state $s'$
with the same feature valuation $f=f(s')$ that is reachable in $P$ in
less number of steps than $s$, the Markovian property is defined as follows:

\begin{definition}[Markovian]
  \label{def:markovian}
  A policy $\pi_\Phi$ is \textbf{Markovian} for a problem $P$ iff the existence of a
  transition $(s,s')$ compatible with $\pi_\Phi$ with feature valuations $f=f(s)$ and $f'=f(s')$,
  implies the existence of transitions $(s_1,s'_1)$ with the same feature valuations $f=f(s_1)$ and $f'=f(s'_1)$,
  in all states $s_1$ that are optimal for $f$ in $P$.
  The policy is \textbf{Markovian} for a class of problems $\Q$ if it is so for each $P$ in $\Q$.
\end{definition}

The states $s_1$ on which the Markovian condition is required in the definition
are not necessarily among those which are reachable with the policy. 
The definition captures a \emph{weak} Markovian assumption. 
% It requires the same possible feature valuation changes $(f,f')$ in all the states $s_1$ that
% reach the feature valuation $f$ optimally, provided that this change occurs in some state transition $(s,s')$ compatible with the policy.
% In such a case, the state transition $(s_1,s'_1)$ will be compatible with the policy as well, as this compatibility
% is a function of the feature valuations $f(s_1)$ and $f(s'_1)$, which are equal to $f(s)$ and $f(s')$ respectively.
The standard, but stronger, Markovian assumption requires the same condition on all states $s_1$
that reach the feature valuation $f$ and not just those that reach $f$ optimally.
% Such stronger
% Markovian assumption implies the weak one, yet we will need the broader definition.
A sufficient condition that ensures the (strong) Markovian property is for the policy to be
\textbf{deterministic} over the feature valuations, meaning that if $f$ is followed by $f'$
in some transition compatible with the policy, $f$ can always be followed by $f'$ in every state
transition $(s,s')$ compatible with the policy where $f(s)=f$, and moreover, that
$f$ can only be followed by $f'$ then. This form of determinism results when the bodies $C$ of the policy rules
$C \mapsto E$ are logically inconsistent with each other, the heads $E$
do not have uncertain effects $\UNK{p}$ or $\UNK{n}$, the domain is
such that all increments and decrements $\INC{n}$ and $\DEC{n}$ are by
fixed amounts, and there is a transition $(s,s')$ compatible with $E$
in the reachable states where $C$ holds.

For reasoning about the policy by reasoning about the features,
the features however must also distinguish goal from non-goal states:

\begin{definition}[Separation] % BLAI: commented to fit def in less lines
  \label{def:sep}
  The features $\Phi$ \textbf{separate goals from non-goals} in $\Q$
  iff there is a set of boolean feature valuations $\kappa$ such that
  for any problem $P$ in $\Q$ and any \emph{reachable} state $s$ in $P$,
  $s$ is a goal state iff $f(s)$ is in $\kappa$. %$s$ determines a boolean feature valuation in $\kappa$.
  The valuations in $\kappa$ are called \textbf{goal valuations}.
\end{definition}

The \textbf{boolean feature valuations} determined by state $s$
refer to the truth valuations of the expressions $p$ and $n=0$ for the
boolean and numerical features $p$ and $n$ in $\Phi$, respectively.
%over the expressions $p(s)=true$ and $n(s)=0$
%that result from a feature valuation over the features $p$ and $n$ in $\Phi$.
%Notice that
While the number of feature valuations is not bounded
as the size of the instances in $\Q$ is not bounded in general,
%as when there is no bound on the size of the instances in $\Q$,
the number of boolean feature valuations is always $2^{|\Phi|}$.

\Omit{
  Sufficient conditions for determinism are simple and common: $\pi_\Phi$ is deterministic in $\Q$
  if it does not include uncertain effects $\UNK{p}$ and $\UNK{n}$, and increments and decrements
  are by constant amounts, e.g.\ $1$. A deterministic policy $\pi_\Phi$ may give rise to different
  plans over an instance $P$, yet in all these plans, the sequence of features values $f_0, \ldots, f_n$
  is the same.
} % omit

The last notion needed for relating general policies and the width of the instances
is the notion of optimality:

\begin{definition}[Optimal policies]
  \label{def:optimal}
  A policy $\pi_\Phi$ that solves a class of problems $\Q$ is \textbf{optimal} if any plan
  $\rho$ induced by $\pi_\Phi$ over a problem $P$ in $\Q$ is optimal for $P$.
\end{definition}

If the policy $\pi_\Phi$ solves a problem $P$, the plans induced by the policy are the action sequences
$\rho$ that yield the goal-reaching state trajectories $s_0, \ldots, s_n$ that are compatible with $\pi_\Phi$.
These plans are optimal for $P$ if there are no shorter plans for $P$.

It can be shown that if $\pi_\Phi$ is a Markovian policy that solves the class of
problems $\Q$ optimally with the features $\Phi$ separating goals from non-goals,
then any feature valuation $f$ reached in the way to the goal in an instance $P$
in $\Q$ is reached optimally; i.e., no action sequence in $P$ can reach a state $s$ with the same feature valuation $f(s)=f$
in a smaller number of steps.

\begin{theorem}
  \label{lemma:optimal}
  Let $\pi_\Phi$ be a \textbf{Markovian} policy that {solves} a class of problems
  $\Q$ \textbf{optimally,} and where the features $\Phi$ \textbf{separate goals} in $\Q$.
  Then, $\pi_\Phi$ is optimal relative to feature valuations.
  That is, if $\rho$ is a sequence of actions induced by $\pi_\Phi$ over
  an instance $P \in \Q$ that reaches the state $s_i$, $\rho$ is an
  \textbf{optimal plan} in $P$ for the feature valuation $f(s_i$).
\end{theorem}
\Proof{
  % \emph{The proof here is slightly more complex than necessary as it assumes a
  % weaker Markovian property that proves to be useful and is introduced later on.}
  Let $\tau^*=(s_0,\ldots,s_n)$ be an optimal trajectory for $P$ compatible
  with $\pi_\Phi$, and let $\tau$ be an optimal trajectory that ends in state $s$
  with $f(s)=f(s_i)$, for $0\leq i\leq n$. We want to show that $|\tau|=i$.
  By the Markovian property, there is a $\pi_\Phi$-transition $(s,s')$ with $f(s')=f(s_{i+1})$.
  If $\tau$ extended with the state $s'$ is an optimal trajectory for $f(s_{i+1})$,
  we can extend it again with a $\pi_\Phi$-$(s',s'')$ into a trajectory for $f(s_{i+2})$.
  Otherwise, there is an optimal trajectory $\tau'$ for $f(s_{i+1})$ which can then
  be extended using the Markovian property into a trajectory for $f(s_{i+2})$.
  Thus, repeating the argument, there is an optimal trajectory $\tau_n$ for $f(s_n)$ of
  length at most $|\tau|+n-i$ (where the length of a trajectory is the number of
  transitions in it).
  On the other hand, since $\pi_\Phi$ is optimal and the features separate the goals,
  $\tau_n$ is a goal-reaching trajectory and thus $n\leq|\tau|+n-i$ which implies
  $i\leq|\tau|$.
  Since $\tau^*$ contains a trajectory that reaches $f(s_i)$ on $i$ steps,
  $|\tau|\leq i$ and thus $|\tau|=i$.
}

As a result, under these conditions, the width of the instances in $\Q$ can be
bounded by the number of features $|\Phi|$ in the policy $\pi_\Phi$, provided
that the features are represented explicitly in the instances:

\begin{theorem}
  \label{thm:markovian:width}
  Let $\pi_\Phi$ be a \textbf{Markovian} policy that {solves} a class of problems
  $\Q$ \textbf{optimally,} where the features $\Phi$ \textbf{separate goals}.
  If the features in $\Phi$ are \textbf{explicitly represented} in the instances $P$ in $\Q$,
  $w(P) \leq |\Phi|$.
  % Then, either the instances $P$ in $\Q$ have $w(P)\leq k$, or %\textbf{width bounded by} $k=|\Phi|$, or
  % the representation of the instances can be extended with some of the $k$ features
  % in $\Phi$ such that their extended representations $P'$ have width $w(P') \leq k$.
\end{theorem}
\Proof{
  Let $k=|\Phi|$, let $\tau^*=(s_0,\ldots,s_n)$ be a goal-reaching $\pi_\Phi$-trajectory
  in $P$, and let $t_i=f(s_i)$, $i=0,1,\ldots,n$.
  Clearly, $|t_i|=k$ and we only need to show that the chain $\theta=(t_0,t_1,\ldots,t_{n})$
  is admissible; i.e., it satisfies the 3 conditions in Definition~\ref{def:width}.
  The first condition is direct since $t_0=f(s_0)$.

  For the second condition, let $\rho$ be an optimal plan that achieves tuple $t_i$ at state $s$.
  We need to show that there is an action $a$ in $P$ such that $(\rho,a)$ is an optimal plan for $t_{i+1}$.
  Since the transition $(s_i,s_{i+1})$ is reachable by $\pi_\Phi$ (i.e., belongs to $\tau^*$),
  the Markovian property implies there must be a $\pi_\Phi$-transition $(s,s')$ associated to an
  action $a$ such that $f(s')=f(s_{i+1})=t_{i+1}$.
  By Theorem~\ref{lemma:optimal}, $|\rho|=i$ and any optimal plan for $t_{i+1}$ is of length $i+1$.
  Therefore, the plan $(\rho,a)$ is a plan for $t_{i+1}$ of length $i+1$ that is optimal.

  For the last condition, we show that any optimal plan that achieves $t_n$
  is an optimal plan for problem $P$.
  This is direct since $t_n$ is a goal valuation as $\Phi$ separates goals:
  a trajectory is goal reaching iff it ends is a state that makes $t_n$ true.
  %and there is no shorter plan for $P$ as the policy $\pi_\Phi$ is optimal.
}

Indeed, if a policy $\pi_\Phi$ solves $\Q$ optimally under the given conditions, an
\emph{admissible chain} $t_0,t_1,\ldots, t_n$ of size $k=|\Phi|$ can be formed
for solving each problem $P$ in $\Q$ optimally, where $t_i$ is the valuation
of the features in $\Phi$ at the $i$-th state of any state trajectory that
results from applying the policy.
% If the features in $\Phi$ are not part of the language, they can be added,
% but often this is not needed.
%but in many cases, this is not needed.

Related to this result, if we let \iwf{\Phi} refer to the variant of IW that replaces tuples of atoms
by feature valuations and thus deems a state $s$ novel in the search (unpruned) when $f(s)$ has not been seen before,
one can show:

\begin{theorem}
  \label{thm:markovian:solutions}
  Let $\pi_\Phi$ be a \textbf{Markovian} policy that \textbf{solves} a class
  of problems $\Q$ \textbf{optimally} with features $\Phi$ that \textbf{separate the goals}.
  The procedure \iwf{\Phi} solves any instance $P$ in $\Q$ \textbf{optimally} in
  time $O(N^{|\Phi|})$ where $N$ is the number of atoms in $P$.
\end{theorem}

\iwf{\Phi} can be thought as a standard breadth-first search that treats
states with the same feature valuations as ``duplicate'' states. It runs
in time $O(N^{|\Phi|})$ as the number of features in $\Phi$ is fixed
and does not grow with the instance size, as opposed to $N$
that stands for the number of atoms in the instance.

\Proof{
  Let $s_0,s_1,\ldots,s_n$ be a goal-reaching trajectory generated by $\pi_\Phi$
  on problem $P$ in $\Q$, and let $f_0,f_1,\ldots,f_n$ be the sequence of feature
  valuations for the states in the trajectory.
  %% In particular, $f_n$ is a goal valuation since the features in $\Phi$ separate the goals.
  %% For simplicity, let us assume that we run \iwf{\Phi} until termination without
  %% stopping it when a goal state is found.
  We prove by induction on $k$ that \iwf{\Phi}
  finds a node $n'_k$ for state $s'_k$ such that $f(s'_k)=f_k$, the path from the root
  node $n_0$ to $n'_k$ is \textbf{optimal}, and the node $n'_k$ is not pruned.
  The base case $k=0$ is direct since the empty path leads to $s_0$ and it is optimal.
  Let us assume that the claim holds for $0\leq k<n$.
  By inductive hypothesis, \iwf{\Phi} finds a node $n'_k$ for state $s'_k$ such
  that $f(s'_k)=f_k$, the path from $n_0$ to $n'_k$ is optimal, and $n'_k$ is not pruned.
  Since the transition $(s_k,s_{k+1})$ exists in $P$, the Markovian property implies
  that there is a transition $(s'_k,s'_{k+1})$ with $f(s'_{k+1})=f_{k+1}$, and thus
  \iwf{\Phi} generates a node $n'_{k+1}$ for state $s'_{k+1}$.
  %%
  %% Moreover, since the optimal cost of reaching $f_{k+1}$ is $k+1$ (Theorem~\ref{lemma:optimal})
  %% there is one such transition leading to a node $n'_{k+1}$ that makes the feature valuation $f_{k+1}$ true.
  If $n'_{k+1}$ is the first node where $f_{k+1}$ holds, it is not pruned.
  Otherwise, there is another node $n''$ that makes $f_{k+1}$ true and the
  length of the path from $n_0$ to $n''$ is less than or equal to the length of
  the path from $n_0$ to $n'_{k+1}$.
  In either case, since the length of an optimal path that achieves $f_{k+1}$ is
  $k+1$ by Theorem~\ref{lemma:optimal}, \iwf{\Phi} finds an optimal path
  to a node that achieves $f_{k+1}$ and is not pruned.

  We have just shown that \iwf{\Phi} finds an optimal path to a state that makes $f_n$ true.
  Since the features separate the goals, such a path is an optimal path for $P$.
}

\begin{example}
  A general policy for Boxes is given by the rules: $\prule{\GT{m}}{\DEC{m}}$ and $\prule{\EQ{m},\GT{n}}{\DEC{n},\UNK{m}}$
  where $m$ and $n$ are two features that count the number of marbles in a box with a least number of marbles, and the
  number of boxes left on the table. Since the policy complies with the conditions in Theorem~\ref{thm:markovian:width}
  and the two features are represented explicitly in the language $\LL_B^4$, it follows that the width of instances of Boxes in such
  encoding is $2$.
\end{example}

\smallskip

\begin{example}
  Similarly, the policy $\pi_\Phi$ with features $\Phi=\{H,n\}$ for $\Q_{clear}$ given
  by the two rules in \eqref{pi:clear} is Markovian, solves $\Q_{clear}$ optimally, and
  the features separate goals. Yet there are no atoms representing the counter $n$ explicitly.
  Still, Theorem~\ref{thm:markovian:solutions} implies that \iwf{\Phi} solves the instances in
  $\Q_{clear}$ optimally in quadratic time.
  % hm:markovian:width} implies that $w(P) \leq 2$ for any $P \in \Q$ if the two
  % features $H$ and $n$ are explicit part of the encoding of $P$ or are captured by suitable atoms.
  %%A more general result that only requires the atoms in $P$ to capture a suitable \emph{projection}
  %%of features is given below.
\end{example}

\smallskip

\Omit{
  Theorem~\ref{thm:markovian:width} can actually be generalized by replacing the assumption of \textbf{determinism}
  by the more general notion of \textbf{optimality} relative to the features $\Phi$; namely,
  a policy $\pi_\Phi$ is optimal with respect to $\Phi$ if for any problem $P \in \Q$,
  any feature valuation $f$ reached by the policy in the way to the goal of $P$
  is reached in a minimum number of steps. Determinism implies optimality relative to
  the features but not the other way around. We skip the formal statement of the theorem
  but illustrates it on an example.
} % omit

Theorem~\ref{thm:markovian:width} relates the number of features in a general policy $\pi_\Phi$ that solves $\Q$
with the width of $\Q$ provided that the features are part of the problem encodings.
This, however, is not strictly necessary:

\begin{theorem}
  \label{thm:width}
  Let $\pi_\Phi$ be an \textbf{optimal} and \textbf{Markovian} policy that solves a class
  $\Q$ of problems over some domain $D$ for which $\Phi$ \textbf{separates the goals}.
  The width of the problems $P$ is bounded by $k$ if for any sequence of feature valuations $\{f_i\}_i$
  generated by the policy $\pi_\Phi$ in the way to the goal, there is a sequence of sets of atoms $\{t_i\}_i$
  in $P$ of size at most $k$ such that the optimal plans for $t_i$ and the optimal plans for $f_i$
  coincide.
  % 1)~the optimal plans for $t_i$ are optimal plans for $f_i$, and
  % 2)~the optimal plans for $f_{i+1}$ that extend those that are optimal for $t_i$, are optimal plans for $t_{i+1}$.
\end{theorem}
\Proof{
  Let $f_0, \ldots, f_n$ be the sequence of feature valuations generated by the policy
  $\pi_\Phi$ in the way to the goal in some instance $P$ in $\Q$, and let $t_0, \ldots, t_n$ be the sequence
  of sets of atoms that comply with the conditions in the theorem, where $|t_i|\leq k$ for $i=0,\ldots,n$.
  We show that $t_0,\ldots,t_n$ is an admissible chain for $P$.

  First, since any optimal plan for $f_0$ is an optimal plan for $t_0$ and
  the empty plan is optimal plan for $f_0$, $t_0\subseteq s_0$.
  Second, if $\rho$ is an optimal plan for $t_n$, $\rho$ is an optimal plan for $f_n$,
  and therefore an optimal plan for $P$ as the features separate the goals.
  Finally, if $\rho$ is an optimal plan for $t_i$ we must show that there is an action $a$
  in $P$ such that $(\rho,a)$ is an optimal plan for $t_{i+1}$, $i < n$.
  Let $\rho$ be an optimal plan for $t_i$ that ends in state $s$.
  Since $\pi_\Phi$ is Markovian and $\rho$ is optimal for $f_i$, there is a transition $(s,s')$
  in $P$ with $f(s')=f_{i+1}$.
  That is, there is an action $b$ such that the plan $(\rho,b)$ reaches $f_{i+1}$.
  By Theorem~\ref{lemma:optimal}, the optimal plans for $f_{i+1}$ are of length $1+|\rho|$,
  and thus $(\rho,b)$ is an optimal plan for $f_{i+1}$
}

\smallskip

\begin{example}
  Consider an instance $P$ in $\Q_{clear}$ where $B_1, \ldots, B_m$ are the blocks
  above $x$ initially, from top to bottom, $m > 0$. The feature valuations in the
  way to the goal following the Markovian policy $\pi_\Phi$ are $f_i = \{H,n=m-i\}$,
  $i=1, \ldots, m$, and $g_i=\{\neg H, n=m-i\}$, $i=0, \ldots, m-1$.
  The policy is optimal, Markovian, and the features separate the goals.
  The tuples of atoms in $P$ that capture these valuations as expressed in
  Theorem~\ref{thm:width} are $t_{f_i}=\{hold(B_i)\}$ and $t_{g_i}=\{ontable(B_i)\}$
  for $i > 0$, and $t_{g_0}=\{clear(B_1)\}$. Since these tuples have size $1$, the
  widths $w(P)$ and $w(\Q_{clear})$ are both equal to $1$.
\end{example}

\section{Admissible Chains and Projected Policies}

Theorem~\ref{thm:width} relates the width of a class $\Q$
to the size of the atom tuples $t_i$ in the instances of $\Q$
that capture the values of the features $f_i$,
following  an optimal  policy  for $\Q$.
We extend this result now by showing that it is often sufficient if the
tuples $t_i$ capture the value of the features $f_i$ in \emph{some} of
those trajectories only.
The motivation is to explain all proofs of bounded width for infinite
classes of problems $\Q$ that we are aware of in terms of general policies.
% The new conditions roughly say that a chain of atom tuples $\theta=(t_0, t_1, \ldots, t_n)$
% is admissible if and only if the tuples $t_i$ capture the feature valuations $f_i$ in \emph{some}
% optimal trajectories compatible with the policy $\pi_\Phi$.
For this, we start with the notion of \emph{feasible chains:}

\begin{definition}[Feasible chain]
  \label{def:theta:feasible}
  Let $\theta=(t_0, t_1, \ldots, t_n)$ be a chain of tuples of atoms from $P$.
  The chain $\theta$ is \textbf{feasible} in problem $P$ if $t_0$ is true in the initial state,
  the optimal plans for $t_n$ have length $n$, and they are all optimal for $P$.
  %% The chain $\theta$ is \textbf{feasible} in problem $P$ if
  %% only $t_0$ holds in the initial state of $P$,
  %% any optimal plan for $t_n$ is optimal for $P$, and
  %% the optimal plans for $P$ have length $n$.
\end{definition}

An \textbf{admissible} chain, as used in the definition of width, is a feasible chain
that satisfies an extra condition; namely, that every optimal plan $\rho$ for
the tuple $t_i$ in the chain can be extended with a single action
into an optimal plan for the tuple $t_{i+1}$, $i=0,1,\ldots,n-1$.
In order to account for this key property, we map feasible chains into
features and policies as follows:

\begin{definition}
  \label{def:policy:theta}
  %For a chain $\theta=(t_0, t_1, \ldots, t_n)$,
  Let $\theta=(t_0, t_1, \ldots, t_n)$ be a chain of atom tuples from $P$, and
  let $\tilde t_i(s)$ denote the \textbf{boolean state feature} that is true in $s$ when
  $t_i$ is true in $s$ and $t_j$ is false for all $i < j \leq n$, $i=1, \ldots, n$.
  The chain defines a \textbf{policy} $\pi_\theta$ over $P$ with rules
  $\prule{\tilde t_i}{\tilde t_{i+1}, \neg\tilde t_i}$, $i=0, \ldots, n-1$.
  %%% where a state transition $(s,s')$ is compatible with $\pi_\theta$ iff
  %%% $\tilde t_i$ is true in $s$, and $\tilde t_{i+1}$ and $\neg\tilde t_i$
  %%% are both true in $s'$.
\end{definition}

% The set of features $\tilde t_i$ used by the policy $\pi_\theta$ is denoted by $\tilde\theta$.
The first result gives necessary and sufficient conditions for a chain $\theta$ to be admissible.

\Omit{% REMOVED since Markovian now is pre-Markovian
  For this, we need a weaker Markovian property of policies that stipulate the
  existence of a transition $(s_2,s'_2)$ only when $s_2$ is a closest state
  (modulo $f(s_2)$) to the initial state:

  \begin{definition}[pre-Markovian]
    \label{def:pre-markovian}
    A policy $\pi_\Phi$ for problem $P$ is \textbf{pre-Markovian} if
    for any $\pi_\Phi$-reachable transition $(s_1,s'_1)$ and any
    \textbf{$\Phi$-closest} non-goal state $s_2$ with $f(s_2)=f(s_1)$,
    there is a $\pi_\Phi$-transition $(s_2,s'_2)$ with $f(s'_2)=f(s'_1)$.
    A state $s$ is $\Phi$-closest if there is no other state $s'$ strictly
    closer than $s$ to the initial state such that $f(s')=f(s)$.
  \end{definition}

  Clearly, any Markovian policy is pre-Markovian, but the converse is not
  necessarily true. Indeed, the policies $\pi_\theta$ induced by admissible
  chains are pre-Markovian but often not Markovian.
  Additionally, by checking the proofs, it is not difficult to see that the
  Theorems~\ref{lemma:optimal}--\ref{thm:width} remain valid when ``Markovian''
  is replaced by ``pre-Markovian'' in the theorem statements.
  We now use the weaker property to characterize admissible chains:
} % omit

\begin{theorem}
  \label{thm:theta}
  Let $\theta=(t_0,t_1,\ldots,t_n)$ be a chain of tuples for a problem $P$.
  $\theta$ is \textbf{admissible} in $P$ if and only if $\theta$ is \textbf{feasible} and the policy
  $\pi_{\theta}$ solves $P$ \textbf{optimally} and is \textbf{Markovian}.
  %% Let $P$ be a problem, and let $\theta=(t_0,t_1,\ldots,t_n)$ be a chain of tuples of atoms in $P$.
  %% $\theta$ is \textbf{admissible} for $P$ if and only if $\theta$ is \textbf{feasible}, the policy
  %% $\pi_{\theta}$ solves $P$ \textbf{optimally}, and $\pi_\theta$ is \textbf{Markovian} for $P$.
\end{theorem}
\Proof{
  Let $s_0$ be the initial state of problem $P$. As always, the length $|\tau|$ of a (state)
  trajectory $\tau=(s_0,\ldots,s_n)$ is $n$. We first establish some claims:
  \begin{enumerate}[C1.]
    %% CLAIM C1
    \item If $\theta$ is admissible, there is no optimal trajectory $\tau$ for $t_i$ that
      also reaches $t_j$ for $0\leq i<j\leq n$.

      \smallskip
      \emph{Proof.} Any optimal trajectory for $t_i$ can be extended with $j-i$ transitions into an
      optimal trajectory for $t_j$, thus no optimal trajectory for $t_i$ can reach $t_j$, $j>i$.
    %%
    %% CLAIM C2
    \item If $\theta$ is admissible, the optimal trajectories for $t_i$ have length $i$ like
      the optimal trajectories for $\tilde t_i$, $0\leq i\leq n$.

      \smallskip
      \emph{Proof.} Let $\tau$ be an optimal trajectory for $t_i$, and let $\tilde\tau$ be an optimal
      trajectory for $\tilde t_i$.
      By definition, $|\tau|\leq|\tilde\tau|$, by C1, $|\tilde\tau|\leq|\tau|$,
      and by admissibility of $\theta$, $|\tau|=i$.
    %%
    %% CLAIM C3
    \item If $\theta$ is feasible, $\pi_\theta$ is optimal and Markovian for $P$, and $\tau$ is a
      trajectory for $t_i$, there is an optimal trajectory for $P$ of length at most $|\tau|+n-i$,
      $0\leq i\leq n$.

      \smallskip
      \emph{Proof.} Let $\tau^*$ be a goal-reaching $\pi_\theta$-trajectory for $P$, and let $\tau$
      be a trajectory for $t_i$. $\tau$ reaches some feature $\tilde t_j$ for $i\leq j\leq n$.
      Then, there is an optimal trajectory $\tau_j$ for $\tilde t_j$, and thus for $t_j$,
      of length $|\tau_j|\leq|\tau|$.
      Using the Markovian property with $\tau^*$, the trajectory
      $\tau_j$ can be extended with a $\pi_\theta$-transition that reaches $\tilde t_{j+1}$.
      Repeating the argument, we find an optimal trajectory $\tau_n$ for $\tilde t_n$ of length
      at most $|\tau|+n-i$.
    %%
    %% CLAIM C4
    \item If $\theta$ is feasible, $\pi_\theta$ is optimal and Markovian for $P$, and $\tau$
      is an optimal trajectory for $t_i$, $|\tau|=i$, $0\leq i\leq n$.

      \smallskip
      \emph{Proof.} Let $\tau$ be an optimal trajectory for $t_i$. By C3, there is an
      optimal trajectory for $t_n$ of length at most $|\tau|+n-i$.
      By feasibility of $\theta$, the length of any optimal trajectory for $t_n$ is exactly $n$.
      Therefore, $n \leq |\tau|+n-i$ which implies $i \leq |\tau|$.
      On the other hand, any goal-reaching trajectory $\tau^*$ induced by $\pi_\theta$
      contains a subtrajectory of length $i$ for $t_i$, and thus $|\tau|\leq i$.
    %%
    %% CLAIM C5
    \item If $\theta$ is feasible, $\pi_\theta$ is optimal and Markovian for $P$, and $\tau$
      is an optimal trajectory for $t_i$, $\tau$ is also optimal for $\tilde t_i$, $0\leq i\leq n$.

      \smallskip
      \emph{Proof.} In the proof of C3, if the feature $\tilde t_j$ is for $j>i$,
      we can find an optimal trajectory for $P$ of length at most $|\tau|+n-j \leq n-1$
      since $|\tau|=i$ by C4. Then, $\tau$ reaches $\tilde t_i$.
      On the other hand, if $\tau'$ is an optimal trajectory for $\tilde t_i$ with
      $|\tau'|<|\tau|$, using the Markovian property we can construct a goal-reaching
      trajectory of length strictly less than $n$.
      Therefore, $\tau$ is optimal for $\tilde t_i$.
  \end{enumerate}

  \textbf{Forward implication.}
  Let us assume that $\theta$ is admissible for $P$. Then, by definition, $\theta$ is \textbf{feasible.}
  Let us now consider a maximal $\pi_\theta$-trajectory $\tau$ in $P$.
  Since the initial state $s_0$ makes true $\tilde t_0$, by admissibility and C1 and C2,
  the trajectory $\tau$ is of length $n$ and ends in a state $s_n$ that makes true $\tilde t_n$.
  In particular, $\tau$ is an optimal trajectory for $t_n$ and thus it is an optimal goal-reaching
  trajectory for $P$.
  This shows that $\pi_\theta$ is \textbf{optimal.}
  Finally, to see that $\pi_\theta$ is Markovian for $P$, let $(s_1,s'_1)$ be a $\pi_\theta$-reachable
  transition and let $s_2$ be a state closest to $s_0$ with $f(s_2)=f(s_1)$. %$\tilde\theta$-closest non-goal state in $P$ with $f(s_1)=f(s_2)$.
  On one hand, By definition, there is one and only one policy rule in
  $\pi_\theta$ that is compatible with $(s_1,s'_1)$, and thus the states
  $s_1$ and $s'_1$ make true the features $\tilde t_i$ and $\tilde t_{i+1}$ respectively.
  On the other hand, by C2, $s_2$ is a closest state to $s_0$ that makes $t_i$ true.
  Hence, by admissibility, the trajectory $\tau$ that leads to $s_2$ can be extended
  with one transition $(s_2,s'_2)$ into an optimal trajectory for $t_{i+1}$.
  Therefore, by C1, the state $s'_2$ makes true $\tilde t_{i+1}$, and the transition
  $(s_2,s'_2)$ is compatible with $\pi_\theta$. Hence, $\pi_\theta$ is \textbf{Markovian.}

  \textbf{Backward implication.}
  Let us assume that $\theta$ is feasible, and that $\pi_\theta$ is optimal and Markovian for $P$.
  We need to show that $\theta$ is an admissible chain; i.e.,
  1)~$t_0$ is true in the initial state $s_0$,
  2)~any optimal plan for $t_i$ can be extended into an optimal plan for $t_{i+1}$ by adding a single action, and
  3)~any optimal plan for $t_n$ is an optimal plan for $P$.
  Conditions 1 and 3 follow directly from the definition of feasible chains.
  For 2, let $\tau=(s_0,\ldots,s_i)$ be an optimal trajectory that reaches $t_i$; its length is $i$ by C4.
  We want to show that $\tau$ can be extended with a transition $(s_i,s_{i+1})$ into an optimal
  trajectory for $t_{i+1}$.
  By C5, $\tau$ is an optimal trajectory for $\tilde t_i$, and thus,
  by the Markovian property, there is a transition $(s_i,s_{i+1})$ that is compatible
  with $\pi_\theta$. Therefore, the state $s_{i+1}$ makes true $\tilde t_{i+1}$ and $t_i$.
  The extended trajectory $\tau'=(s_0,\ldots,s_{i+1})$ is optimal for $t_{i+1}$ by C4.
}

This result connects admissible chains with policies that are optimal and Markovian,
but it does not connect admissible chains with general policies.
This is done next. We first define when a policy $\pi_1$ can be regarded as a
\textbf{projection} of another policy $\pi_2$:

\begin{definition}[Projection]
  \label{def:projection}
  Let $\pi_\Phi$ be a policy over a class $\Q$ and let $\pi_{\Phi'}$ be a policy for a problem $P$ in $\Q$.
  The policy $\pi_{\Phi'}$ is a \textbf{projection} of $\pi_{\Phi}$ in $P$ if every
  \textbf{maximal state trajectory} compatible with $\pi_{\Phi'}$ in $P$ is a \textbf{maximal state trajectory}
  in $P$ compatible with $\pi_{\Phi}$.
\end{definition}

Notice that it is not enough for the state trajectories $s_0, \ldots, s_i$ compatible with $\pi_{\Phi'}$
to be state trajectories compatible with $\pi_\Phi$; it is also required that if $s_i$ is a final state
in the first trajectory that it is also a final state in the second one.
This rules out the possibility that there is a continuation of the first trajectory that is compatible
with $\pi_\Phi$ but not with $\pi_{\Phi'}$. A result of this is that if $\pi_\Phi$
is optimal for $\Q$, the projected policy $\pi_{\Phi'}$ must be optimal for $P$.
From this, the main theorem of this section follows:

\Omit{% Following theorem subsumed
  \begin{theorem}
    \label{thm:projection}
    Let $P$ be a problem, and let $\theta=(t_0,t_1,\ldots,t_n)$ be a chain of atom tuples for $P$.
    $\theta$ is \textbf{admissible} if and only if $\theta$ is \textbf{feasible}, $\pi_\theta$ is a \textbf{projection}
    of an \textbf{optimal} policy $\pi_\Phi$ for $P$, and $\pi_\theta$ is \textbf{pre-Markovian}.
  \end{theorem}
  \Proof{
    If $\theta$ is admissible, by Theorem~\ref{thm:theta}, $\theta$ is feasible,
    and $\pi_\theta$ is pre-Markovian and solves $P$ optimally. On the other hand,
    $\pi_\theta$ is a projection of itself.

    If $\theta$ is feasible, $\pi_\theta$ is pre-Markovian and it is the projection of an optimal
    policy $\pi_\Phi$, then $\pi_\theta$ is optimal, and by the same theorem, $\theta$ is admissible.
    To see that $\pi_\theta$ is optimal, observe that any maximal $\pi_\theta$ trajectory is
    a maximal $\pi_\Phi$ trajectory and thus goal-reaching and optimal.
  }
} % omit

\begin{theorem}
  \label{thm:projection:class}
  Let $\pi_\Phi$ be an \textbf{optimal} policy for a class $\Q$ of problems.
  If for any problem $P$ in $\Q$, there is a \textbf{feasible} chain $\theta$
  of size at most $k$ such that $\pi_\theta$ is a \textbf{projection} of $\pi_\Phi$ in $P$
  that is \textbf{Markovian}, then $w(\Q)\leq k$.
\end{theorem}
\Proof{
  Theorem~\ref{thm:theta} implies that if $\theta$ is feasible and $\pi_\theta$ is Markovian
  and a projection of an optimal policy $\pi_\Phi$ in $P \in \Q$, then $\theta$ is admissible,
  and hence that $w(P)$ is bounded by the size of $\theta$ (i.e., max size of a tuple in $\theta$).
  If this size is bounded by $k$ for all $P$ in $\Q$, $w(\Q)$ is bounded by $k$.
}

While the proof of this theorem is a direct consequence of Theorem~\ref{thm:theta}, the result
is important as \emph{it renders explicit the logic underlying all proofs of bounded width for
meaningful classes of problems $\Q$ that we are aware of.}
In all such cases, the proofs have been constructed by finding feasible chains with tuples
$t_i$ that are a function of the instance $P \in \Q$, and whose role is to capture suitable {projections}
of \textbf{some general optimal policy} $\pi_\Phi$.

\Omit{
  % In other words, while the theorem goes in one direction,
  % from optimal policies for a class $\Q$ to the width of $\Q$, the significance of the theorem goes in the other direction,
  % that cannot be proved and can only be conjectured; namely, that proofs of bounded width for meaningful, infinite classes
  % of problems $\Q$ all start with a general optimal policy. The reason that the converse of the theorem cannot be proved
  % is that we are not demanding any structure on $\Q$.
  % , and nothing prevents from $\Q$ being made of a single instance $P$
  % whose width is not structural but accidental. The theorem does not put conditions on $\Q$ but on the policies that solve $\Q$
  % and their projections.

  It is worth noticing that in Theorems~\ref{thm:markovian:width} and \ref{thm:width}
  the tuples $t_i$ in admissible chains $\theta$ are required to capture the feature valuations $f_i$ along
  \textbf{all} trajectories compatible with an optimal policy $\pi_\Phi$, in
  Theorem~\ref{thm:projection:class},
  the tuples $t_i$ are required to capture the feature valuations $f_i$
  along \textbf{some} of such trajectories only; namely, those of a projection of $\pi_\Phi$.
} % omit

\Omit{
  those generated by the policy
  $\pi_\theta$ determined by $\theta$, provided that $\pi_\theta$ is a projection
  of a general optimal policy.
  If the conditions in the theorem hold, the tuples in the chain are said to be a
  \textbf{projection} of the features $\Phi$ for the policy $\pi_\Phi$ on the
  instance $P$.
  %% The theorem provides a \textbf{general explanation for why the width of a class of
  %% problems $\Q$ is often bounded:} it is because tuples of atoms of bounded size in
  %% each of the problems in $\Q$ capture suitable projections of the features for an
  %% optimal policy for $\Q$.

  On the other hand, these theorems provide an effective handle to bound
  the width $w(\Q)$ for classes of problems.
  Indeed, since the features in $\tilde\theta$ that define the policy rules
  in $\pi_\theta$ are boolean, and such rules have uncertain effects (i.e., effects
  with question marks) and their conditions are pairwise inconsistent, then
  $\pi_\theta$ is pre-Markovian iff for any $\pi_\theta$-reachable transition
  $(s_1,s'_1)$, and $\tilde\theta$-closest state $s_2$ with $f(s_1)=f(s_2)$,
  there is a $\pi_\theta$-transition $(s_2,s'_2)$.
} % omit

\smallskip

\begin{example}
  Let $\Q_G$ be a grid navigation domain where an agent at some initial cell has to move
  to a target cell with atoms of the form $x(i)$ and $y(j)$ that specify the coordinates
  of the agent along the two axes. An optimal policy for $\Q_G$ can be obtained with
  the singleton $\Phi=\{d\}$ where the linear feature $d$ measures the distance to the target cell,
  $\EQ{d}$ identifies the goal states, and there is a
  single policy rule $\prule{\GT{d}}{\DEC{d}}$. %Clearly, $\EQ{d}$ identifies the goal states.
  Let $P$ be a problem in $\Q_G$ where the agent is initially located at cell $(i_0,j_0)$ and the goal cell is $(i_n,j_m)$,
  which for simplicity satisfies $i_0\leq i_n$ and $j_0\leq j_m$.
  Further, let $\theta=(t_0,t_1,\ldots,t_{n+m})$ be the chain of tuples where
  $t_k=\{x(i_0+k)\}$ for $0\leq k< n$, and $t_k=\{x(i_n),y(j_0+k-n)\}$ for $n\leq k\leq m+n$.
  Then, since $\theta$ is feasible, and $\pi_\theta$ is the projection of the optimal policy
  $\pi_\Phi$ while being Markovian, it follows from Theorem~\ref{thm:projection:class} that $w(\Q_G)\leq 2$.
  Notice that the tuples $t_i$ in $\theta$ do not capture the values $d_i$ of the distance feature $d$
  in \textbf{all} the trajectories that are compatible with the policy $\pi_\Phi$ in $P$ (an exponential number of such trajectories),
  but in \textbf{one} such trajectory only; namely, the one that is compatible with the projection $\pi_\theta$ of $\pi_\Phi$,
  with states $s_i$, $i=0, \ldots, n+m$, where the tuple $t_i$ is true iff the feature $\tilde t_i$ used in $\pi_\theta$ is true.
\end{example}

\smallskip

\begin{example}
  In the Delivery ($D$) domain an agent moves in a grid to pick up packages and
  deliver them to a target cell, one by one; {Delivery-1 ($D_1$)} is the version
  with one package.
  A general policy $\pi_\Phi$ for $D$ and $D_1$ is given in terms of four rules
  that capture: move to the (nearest) package, pick it up, move to the target cell,
  and drop the package, in a cycle, until no more packages are left.
  The rules can be expressed in terms of the features $\Phi=\{H,p,t,n\}$ that express
  holding, distance to the nearest package (zero if agent is holding a package or no
  package to be delivered remains), distance to the target cell, and the number of
  undelivered packages respectively. Hence, $\EQ{n}$ identifies the goal states for
  the problems in the classes $\Q_D$ and $\Q_{D_1}$ for the problems $D$ and $D_1$
  respectively.
  The rules that define the policy $\pi_\Phi$ are $\prule{\neg H,\GT{p}}{\DEC{p},\UNK{t}}$,
  $\prule{\neg H, \EQ{p}}{H}$, $\prule{H,\GT{t}}{\DEC{t}}$, and
  $\prule{H,\GT{n},\EQ{t}}{\neg H, \DEC{n}, \UNK{p}}$.\footnote{A different formulation
    involves packages that need to be delivered to target cells that depend on the package.
    The set of features is the same $\Phi=\{H,p,t,n\}$ except that $t$ is defined as the
    distance to the \emph{current target cell} (zero if agent holds nothing or there are
    no more packages to deliver).
    A policy for this formulation has the rules $\prule{\neg H, \GT{p}}{\DEC{p}}$,
    $\prule{\neg H, \EQ{p}}{H, \INC{t}}$, $\prule{H, \GT{t}}{\DEC{t}}$, and
    $\prule{H, \GT{n}, \EQ{t}}{\neg H, \DEC{n}, \UNK{p}}$.
  }

  Let us consider a problem $P$ in the class $\Q_{D_1}$.
  If the encoding of $P$ contains atoms like $at(cell_i)$, $atp(pkg_i,cell_j)$, $hold(pkg_i)$,
  and $empty$, it can be shown that $\Q_{D_1}$ has width $2$.
  Indeed, without loss of generality, let us assume that in $P$, the package is initially
  at $cell_i$ and has to be delivered at $cell_t$, and the agent is initially at $cell_0$,
  and let $\theta=(t_0, t_1, \ldots, t_n)$ be the chain made of tuples of the form $\{at(cell_k)\}$
  for the cells on a shortest path from $cell_0$ to $cell_i$, followed by tuples of the form
  $\{at(cell_k),hold(pkg_j)\}$, with $cell_k$ now ranging over a shortest path from $cell_i$
  up to $cell_t$, and a last tuple $t_n$ of the form $\{at(pkg_j,cell_t)\}$.
  It is easy to show that the chain $\theta$ is feasible, and $\pi_\theta$ is the projection of the general policy $\pi_\Phi$
  that in $D_1$ is both optimal and Markovian (although not in $D$).
  Then, by Theorems~\ref{thm:theta} and \ref{thm:projection:class}, it follows
  that the chain $\theta$ is admissible, and %since the same argument holds for variations of $P$, that
  $w(\Q_{D_1})\leq 2$.
\end{example}

\section{Serialized Width}

Theorem~\ref{thm:projection:class} suggests that bounded width
is most often the result of general policies that can be
projected on suitable tuples of atoms. We extend now these
relations to the much larger and interesting class of problems that have
bounded \emph{serialized width}, where the general policies do not have to
be optimal or Markovian. The notion of {serialized width} is based on the decomposition
of problems into subproblems, and explicates and generalizes the logic and scope
of the Serialized IW (SIW) algorithm of \citeay{nir:ecai2012}.
There are indeed many domains with large or unbounded widths that are
simple and admit simple policies, like the Delivery domain above.

In SIW, problems are serialized by using one feature, a \emph{goal counter}
denoted by $\#g$, that tracks the number of unachieved goal atoms, assuming
that goals are conjunctions of atoms.
Other serializations have appealed to other counters and also to heuristics
\cite{nir:aaai2017}.\footnote{These serializations appear in state-of-the-art
  algorithms like BFWS that use novelty measures inside complete best-first
  procedures.
}

We take a more general and abstract approach that replaces the goal
counter by a \emph{strict partial order} (an irreflexive and transitive
binary relation), over the feature valuations, also referred to as $\Phi$-tuples.
A serialization for a class $\Q$ of problems is defined as follows:

\begin{definition}[Serializations]
  \label{def:serialization}
  Let $\Q$ be a class of problems, let $\Phi$ be a set of goal-separating features
  for $\Q$, and let $\prec$ be a \textbf{strict partial order} over $\Phi$-tuples.
  The pair $(\Phi,\prec)$ is a \textbf{serialization} over $\Q$ if
  1)~the ordering $\prec$ is well-founded; i.e.\ there is no infinite descending chain
  $f_1 \succ f_2 \succ f_3 \succ \cdots$ where $f_i \succ f_{i+1}$ stands for
  $f_{i+1} \prec f_i$, and 2)~the goal feature valuations $f$ are $\prec$-minimal;
  i.e., no $f' \prec f$ for any feature valuation $f'$.
  %%% \begin{enumerate}[1.]
  %%%   \item $\prec$ is well founded (no infinite descending chain), and
  %%%   \item the goal feature valuations are $\prec$-minimal.
  %%% \end{enumerate}
  %%% In particular, conditions 2--3 are satisfied when $f(s)$ is $\prec$-minimal iff $s$ is a goal state.
  %%% $(\Phi,\prec)$ is a serialization for $\Q$ iff it is a serialization for any $P$ in $\Q$.
  %%% $(\Phi,\prec)$ is a serialization for $\Q$ iff it is a serialization for any $P$ in $\Q$.
  %%%
  %%% The serialization is \textbf{accessible} when the test $t\prec t'$, for tupes $t$ and $t'$,
  %%% can be done in constant time.\footnote{For example, this is satisfied for partial orders
  %%% $\prec$ defined by a lexicographic ordering over tuples since, by the assumption
  %%% on the computational model, comparisons over integers can be done in constant time.}
\end{definition}

A serialization over $\Q$ decomposes each problem $P$ into subproblems which
define the width of the serialization or serialized width:

\begin{definition}[Subproblems]
  \label{def:serialization:subproblems}
  Let $(\Phi,\prec)$ be a serialization. The subproblem $P[s,\prec]$ is the problem of
  finding a state $s'$ reachable from $s$ such that $s'$ is goal in $P$ or $f(s')\prec f(s)$;
  i.e., the goal states in $P[s,\prec]$ are the goal states in $P$ and the states $s'$ such that $f(s')\prec f(s)$.
  The collection $P[\prec]$ of subproblems for problem $P$ is the smallest subset
  of problems $P[s,\prec]$ that comply with:
  \begin{enumerate}[1.]
    \item if the initial state $s_0$ in $P$ is not goal, $P[s_0,\prec] \in P[\prec]$, % for the initial state $s_0$ in $P$,
    \item $P[s',\prec] \in P[\prec]$ if $P[s,\prec] \in P[\prec]$ and $s'$ is a non-goal state in $P$ that is at
      \textbf{shortest distance} from $s$ such that $f(s')\prec f(s)$, and no goal state of $P$ is strictly closer from $s$ than $s'$
      %%distance $d'\leq d$ from $s$ and no state $s''$ with $f(s'')=f(s')$ is at
      %%distance $d'<d$ from $s$.
  \end{enumerate}
\end{definition}

%%\noindent The serialized width of a problem is the max width of its subproblems:

\begin{definition}[Serialized width]
  \label{def::serialization:width}
  Let $(\Phi,\prec)$ be a serialization for a collection $\Q$ of problems. %, and let $k$ be a non-negative integer.
  The \textbf{serialized width} of problem $P$ relative to $(\Phi,\prec)$ is $k$, written as $w_\Phi(P)=k$
  with the ordering ``$\prec$'' left implicit, if there is a subproblem $P[s,\prec]$ in $P[\prec]$ that has width $k$,
  and every other subproblem in $P[\prec]$ has width at most $k$.
  %if $P[\prec]$ is empty, $w_\Phi(P)$ is defined as 0.
  The \textbf{serialized width} for $\Q$ is $k$, written as $w_\Phi(\Q)=k$, if $w_\Phi(P)=k$ for some
  problem $P\in \Q$, and $w_\Phi(P)\leq k$ every other problem $P'\in \Q$.
\end{definition}

If a class of problems $\Q$ has bounded serialized width and the ordering
$f\prec f'$ can be tested in polynomial time, each problem $P$ in $\Q$
can be solved in polynomial time using a variant \siw of the SIW
algorithm: starting at the state $s=s_0$, \siw performs an IW search
from $s$ to find a state $s'$ that is a goal state or that renders
the precedence constraint $f(s')\prec f(s)$ true.
If $s'$ is not a goal state, $s$ is set to $s'$, $s:=s'$, and
the loop repeats until a goal state is reached.
The result below follows from the observations by \citeay{nir:ecai2012} once the
goal counter is replaced by a partial order, and the notion of serialized
width is suitably formalized:
%%If the ordering $f \prec f'$ can be tested in polynomial time,
%%it follows that a bounded serialized width for a class of problems $\Q$,
%%implies that the problems $P$ in $\Q$ can be solved in polynomial time,
%%using a variant SIW$_\Phi$ of the SIW algorithm, that starting with $s=s_0$, performs
%%an IW search from $s$ to a state $s'$ such that $f(s')\prec f(s)$ or $s'$ is a goal state.
%%If $s'$ is not a goal state, $s$ is set to $s'$, $s:=s'$, and the loop repeats until a goal state
%%is reached. The result follows from the observations in \cite{nir:ecai2012}
%%once the goal counter is replaced by partial order, and the notion of serialized width is suitably formalized:

\begin{theorem}
  \label{thm:serialization:siw}
  Let $\Q$ be a collection of problems and let $(\Phi,\prec)$ be
  a serialization for $\Q$ with width $w_\Phi(\Q)\leq k$.
  %If the serialization order $f \prec f'$ can be tested in \emph{constant time,}
  Any problem $P$ in $\Q$ is solved by SIW$_\Phi$ in time
  and space bounded by $O(bN^{|\Phi|+2k-1}\Lambda)$ and $O(bN^k + N^{|\Phi|+k})$
  respectively, where $b$ bounds the branching factor in $P$,
  $N$ is the number of atoms in $P$, and $\Lambda$ bounds the
  time to test the order $\prec$ for the $\Phi$-tuples that arise from
  the states in $P$.
\end{theorem}
\Proof{
  We show something more general.
  If $P$ is a problem in $\Q$, \siw finds a plan $\rho$ for $P$
  and state sequence $s_0,s_1,\ldots,s_n$ such that
  \begin{enumerate}[1.]
    \item the state $s_0$ is the initial state in $P$,
    \item the state $s_{i+1}$, reachable from state $s_i$, can be found by \iw{k} on the problem $P[s_i,\prec]$, for $i=0,1,\ldots,n-1$,
    \item $f(s_{i+1}) \prec f(s_i)$, for $i=0,1,\ldots,n-1$,
    \item $s_n$ is a goal state, and
    \item $|\rho|\leq N^{|\Phi|+k}$.
  \end{enumerate}
  \siw achieves this in time and space bounded by $O(bN^{|\Phi|+2k-1} \Lambda)$ and $O(bN^k + N^{|\Phi|+k})$ respectively.

  By definition, \siw calls IW iteratively to find such a sequence.
  In the first call, over subproblem $P[s_0,\prec]$, IW finds a state $s_1$ such that
  $f(s_1)\prec f(s_0)$. Then, iteratively, after $s_i$ is found, IW is called to solve
  the subproblem $P[s_i,\prec]$. The process continues until IW finds a goal state $s_n$.

  Let us assume that such a sequence is found by \siw.
  A run of IW involves runs of $\iw{1}, \iw{2}, \ldots$ until the problem is solved
  (guaranteed to succeed before the call to \iw{k+1} since $w_\Phi(\Q)\leq k$).
  By Theorem~\ref{thm:width:nir} and the assumption on the cost of testing $\prec$, each
  call to IW requires time and space bounded by $O(bN^{2k-1}\Lambda)$ and $O(bN^{k})$ respectively.
  The plan $\rho_i$ found by IW that maps $s_i$ into $s_{i+1}$ has length bounded by $N^k$.
  All these plans are concatenated into a single plan $\rho$ that solves $P$.
  Since the number of $\Phi$-tuples is bounded by $N^{|\Phi|}$,
  \siw invests total time and space bounded by $O(bN^{|\Phi|+2k-1})$ and $O(bN^k + N^{|\Phi|+k})$
  respectively, and $|\rho|\leq N^{|\Phi|+k}$.

  We now show that the state sequence is indeed found.
  The first requirement is clear since $s_0$ is the initial state in $P$.
  We reason inductively to show that the subproblem $P[s_i,\prec]$ belongs to the collection
  $P[\prec]$ of subproblems, and that the state $s_{i+1}$ is at minimum distance from $s_i$
  such that $f(s_{i+1})\prec f(s_i)$.

  For the base case, $P[s_0,\prec]$ belongs to $P[\prec]$ by definition of $P[\prec]$ when
  $s_0$ is not a goal state; if $s_0$ is a goal state, the sequence with the single state
  $s_0$ satisfies the requirements.
  Since $w_\Phi(P)\leq k$, \iw{k} is guaranteed to find a state $s_1$ at minimum distance
  from $s_0$ that is a goal or $f(s_1)\prec f(s_0)$.
  If $s_1$ is a goal state, the sequence $s_0,s_1$ satisfies the requirements.
  Otherwise, by definition of $P[\prec]$, $P[s_1,\prec]$ belongs to $P[\prec]$.
  Assume now that $s_i$ is not a goal state and $P[s_i,\prec]$ belongs to $P[\prec]$.
  A similar argument shows that \iw{k} finds a state $s_{i+1}$ at
  minimum distance from $s_i$ that is a goal or $f(s_{i+1})\prec f(s_i)$.
  If $s_{i+1}$ is a goal state, \siw has found the required state sequence.
  Otherwise, the problem $P[s_{i+1},\prec]$ belongs to $P[\prec]$.
  Finally, since the number of $\Phi$-tuples is finite, this process
  terminate with a goal state $s_n$.
}

The class $\Q_{V\!A}$ of instances for the problem {VisitAll}, where all cells in a grid must be visited,
has serialized width $w_\Phi(\Q_{V\!A})=1$ for $\Phi=\{\#g\}$ where $\#g$ counts %that has the feature that counts
the number of unvisited cells (unachieved goals), with the obvious ordering ('$\prec$' set to '$<$').
The class $\Q_{BW}$ of {Blocksworld} instances cannot be serialized into subproblems
of bounded width with $\Phi'=\{\#g\}$ because achieved goals may have to be undone,
yet with $\Phi=\{\#m\}$ where $\#m$ counts the number of misplaced blocks (i.e., not on their targets
or above one such block), $w_\Phi(\Q_{BW})=2$.
%Since the number of joint feature valuations for such $\Phi$
%is linear, the result above implies that SIW$_\Phi$ solves these problems in polynomial time.
The result above implies that SIW$_\Phi$ solves these problems in polynomial time.

\section{From General Policies to Serializations}

We show next how serializations can be inferred from general policies,
a result that paves the way to introduce a convenient language for
defining serializations. % We answer two questions in particular: under what
% conditions a general policy $\pi_\Phi$ defines a serialization of bounded width,
% and under what conditions, bounded serialized width, and in particular, serialized width equal to $1$,
% allow us to construct a general policy.
Given a general policy $\pi_\Phi$ that solves a class of problems $\Q$
for features $\Phi$ that separates the goals, the intuition is to consider serializations
$(\Phi,\prec)$ where $f' \prec f$ % for two feature valuations $f$ and $f'$
if there is a state transition $(s,s')$ compatible with the policy $\pi_\Phi$
in some instance $P$ in $\Q$ with $f'=f(s')$ and $f=f(s)$.
The problem with this ordering, however,
is that it does not necessarily define a strict partial order as required, as it is
possible that there is another transition $(s_1,s_2)$ compatible with the policy
in another instance $P' \in \Q$ where $f(s_1)=f'$ and $f(s_2)=f$. In other words,
even if the policy orders the feature valuations in a strict partial order in an instance $P$,
it is possible that the orderings over two or more instances in $\Q$ cannot be combined
into a single strict partial order. \Omit{
  \footnote{For an example of this, consider an agent in a $1 \times N$ row of cells
    and two features $n$ and $m$ in $\Phi$ that count the number of cells
    to the left and to the right of the agent respectively. The goal is to make one of the two features zero; i.e., $n=0$ or $m=0$.
    The policy $\pi_\Phi$ with rules $\prule{}{\DEC{n}, \INC{m}}$ and $\prule{}{\INC{n},\DEC{m}}$, tells the agent to move either
    left or right. Since the agent can interleave the two types of moves, the policy does not solve this class of problems $\Q_{LR}$.
    On the other hand, if $\Q_{LR}^-$ is defined instead as the family of instances where the agent can move in one direction only
    (because moves in the other direction are not allowed), then the policy $\pi_\Phi$ solves both $\Q_{LR}^-$
    but it does not define a strict partial order among feature valuations. Indeed, in the instances in $\Q$ where
    only left moves are allowed and $N=10$, the feature valuation $f=(4,6)$
    is preferred to feature valuation $f'=(5,5)$, but in the other instances,
    it'll be exactly the opposite.
  }
} % omit

Thus in order to obtain a serialization $(\Phi,\prec)$ over a class of problems $\Q$
from a policy $\pi_\Phi$ certain additional restrictions are required. For this,
we focus on the policies $\pi_\Phi$ that solve $\Q$ \textbf{structurally};
i.e., by virtue of the effects of the actions on the features as expressed in the policy rules in $\pi_\Phi$.\Omit{
  \footnote{
    For example, consider the class of problems $\Q_L$ but where the only action allowed is to move
    left and the policy $\pi_\Phi$ with a single feature $\Phi=\{n\}$ that measures the distance to the left most cell,
    expressed by the rule $\prule{\GT{n}}{\DEC{n}}$. This policy solves $\Q_L$ as every transition
    compatible with the policy must decrease $n$ up to $0$. Yet, the policy $\prule{\GT{n}}{\UNK{n}}$ solves $\Q_L$
    as well, because the only possible transitions satisfy this rule too. It's not possible however
    to predict that the second policy will achieve the goal $n=0$ from the form of the policy rules,
    hence, such a policy solves $\Q_L$ by not due its structure.
  }
} % omit
A policy solves a class of problems $\Q$ structurally when this can be verified
from the form of the policy rules without having to consider the instances in $\Q$.
Fortunately, we do not have to formulate the conditions under which a policy $\pi_\Phi$
solves a class of problems $\Q$ structurally from scratch, we can use ideas developed
in the context of QNPs \cite{sid:aaai2011,bonet:qnps}.

The key idea is the notion of \textbf{terminating policy graph.}
Recall that every feature valuation $f$ defines a projected boolean valuation $b(f)$
over the expressions $p$ and $n=0$ for the boolean and numerical features $p$ and $n$ in $\Phi$,
where $p$ and $n=0$ are true in $b(f)$ iff $p$ and $n=0$ are true in $f$ respectively.
The policy graph for policy $\pi_\Phi$ is:

\begin{definition}[Policy graph]
  \label{def:policy:graph}
  The policy graph $G(\pi_\Phi)$ for policy $\pi_\Phi$ has nodes $b$, one for each of the
  $2^{|\Phi|}$ boolean feature valuations over $\Phi$, and edges $b \rightarrow b'$ labeled with $E$
  if $b$ is not a goal valuation and $(b,b')$ is compatible with a rule $C \rightarrow E$ in the policy.
\end{definition}

The edge $b \rightarrow b'$ is compatible with a rule $C \rightarrow E$
if $C$ is true in $b$, and $(b,b')$ is compatible with $E$; namely, if $p$ (resp.\ $\neg p$) is in $E$,
$p$ (resp.\ $\neg p$) must be true in $b'$, and if $\INC{n}$ is in $E$, $\EQ{n}$ must be false in $E$.
Effects $\UNK{p}$, $\UNK{n}$, and $\DEC{n}$ in $E$ put no constraints on $b'$. In particular, in the latter case,
$n=0$ can be either true or false in $b'$, meaning that after a decrement, a numerical feature
may remain positive or have value $0$. Notice that the policy graph associated with a policy $\pi_\Phi$
does not take the class of problems $\Q$ into account. Indeed, the policy graph may have an edge $b \rightarrow b'$ even if there is no state transition $(s,s')$
in an instance in $\Q$ where this transition between boolean feature valuations arises.
The policy graph and the notion of \textbf{termination} below are purely \textbf{structural}
as they depend on the form of the policy rules only:

\begin{definition}[Termination]
  \label{def:termination}
  A policy $\pi_\Phi$ and a policy graph $G(\pi_\Phi)$ are \textbf{terminating} if for every cycle in the graph $G(\pi_\Phi)$, i.e., any
  sequence of edges $b_i \rightarrow b_{i+1}$ with labels $E_i$ that start and end in the same node, there is a numerical feature
  $n$ in $\Phi$ that is decreased along some edge and increased in none.
  That is, $\DEC{n}\in E_k$ for some $k$-th edge in the cycle,
  and $\INC{n}\not\in E_j$ and $\UNK{n}\not\in E_j$ for all others edges in the cycle.
\end{definition}

The termination condition can be checked in time that is polynomial in the size of the policy graph
by a procedure called \textsc{Sieve} that looks at the strongly connected components in the graph
\cite{sid:aaai2011,bonet:qnps}.\footnote{See \citeay{bonet:qnps} for more details on the formal
  properties of termination in QNPs. Our setting is slightly different as our numerical features
  are linear and cannot grow without bound as in QNPs, but we do not use this property to
  define termination.
}
A key property of terminating policies % $\pi_\Phi$ %(over non-negative numerical features)
is that they give rise to state trajectories that are finite. % only, i.e., that have a final state.

\begin{theorem}
  \label{thm:termination}
  Let $\pi_\Phi$ be a terminating policy with features $\Phi$ over $\Q$ that separate the goals.
  Then, $\pi_\Phi$ cannot give rise to infinite state trajectories in instances $P$ in $\Q$.
\end{theorem}
\Proof{
  For a proof by contradiction, let us assume that there is an infinite state trajectory $s_0,s_1,\ldots$ that is compatible with $\pi_\Phi$.
  Since the number of states in $P$ is finite, there must be indices $i<j$ such that $s_i=s_j$.
  %%$s_i, \ldots, s_k$ with $i < k$ such that $s_i=s_k$, and therefore where $f(s_i)=f(s_k)$,
  %%but this is actually impossible if $\pi_\Phi$ is terminating.
  The subsequence $s_i,\ldots,s_j$ defines a cycle $b(f(s_i)),\ldots,b(f(s_j))$ in the policy graph for $\pi_\Phi$.
  The termination condition implies that there is a numerical feature $n$ in $\Phi$
  that is decreased by some edge in the cycle but not increased by any edge in the cycle.
  However, this is impossible since $s_i=s_j$ implies that the feature $n$ has the same value on both states.
  Therefore, such infinite state trajectory cannot exist.
  %%Indeed, if $b_j$ is the boolean valuation
  %%associated with the feature valuation $f(s_j)$, then the policy graph must contain an edge
  %%$b_j \rightarrow b_{j+1}$ for any two consecutive states $(s_j,s_{j+1})$ in the sequence, as the edge
  %%must be compatible with a policy rule if this state transition is compatible with a policy rule.
  %%As a result, the cycle $s_i, \ldots, s_k$ with $i < k$ and $s_i=s_k$, implies the presence
  %%of a cycle $b_i, \ldots, b_k$ in the policy graph, but then termination ensures that
  %%there must be a numerical feature $n$ such that $n$ is decreased in this path and is not increased,
  %%but this implies that $f(s_i)$ and $f(s_k)$ can't be equal, in contradiction with $s_i=s_k$.
}

Since terminating policies induce state trajectories with a final state, %each with a final state,
a sufficient condition for a terminating policy to solve a class $\Q$ is to be closed:
%in order for a terminating policy to solve a class of problem $\Q$, it is enough
%that the policy is closed in the following sense:

\begin{definition}[Closed policy]
  \label{def:policy:closed}
  A policy $\pi_\Phi$ is \textbf{closed} over a class of problems $\Q$ if for any
  non-goal state $s$ in a problem $P$ in $\Q$ that is $\pi_\Phi$-reachable,
  %from the initial state of $P$ with the policy,
  there is a transition $(s,s')$ that is compatible with $\pi_\Phi$.
  %% THE FOLLOWING IS SOUNDNESS
  %% or $s$ is reachable from $s_0$ and the policy $\pi_\Phi$ is \textbf{applicable} at $s$
  %% (i.e., the set of effects $\pi_\Phi(s)$ is non-empty).
\end{definition}

%\noindent As a result:

\begin{theorem}
  \label{thm:structurally}
  If $\pi_\Phi$ is closed over $\Q$, the features in $\Phi$ separate goals
  in $\Q$, and $\pi_\Phi$ is terminating, $\pi_\Phi$ \textbf{solves} $\Q$.
  % The policy $\pi_\Phi$ is said to solve $\Q$ \textbf{structurally.}
\end{theorem}
\Proof{
  If $\pi_\Phi$ does not solve instance $P$ in $\Q$, there is a maximal
  non-goal reaching trajectory in $P$ that is compatible with $\pi_\Phi$.
  This trajectory is either infinite or terminates at state $s_n$ where
  no transition $(s_n,s)$ compatible with $\pi_\Phi$ exists in $P$.
  The former is impossible by Theorem~\ref{thm:termination}, and the
  latter is impossible since $\pi_\Phi$ is closed.
  %% First condition in def of closed policy used here!
}

We are interested in the conditions under which general policies
determine serializations, and in particular, serializations with bounded width.
For the first part, we do not need the policy to be closed or solve $\Q$,
but just to be terminating:

\begin{theorem}
  \label{thm:terminating:serialization}
  A terminating policy $\pi_\Phi$ with features $\Phi$ over $\Q$ that separate goals
  determines a serialization $(\Phi,\prec)$ of $\Q$ where `$\prec$' is
  the minimal strict partial relation (i.e., the transitive closure)
  that satisfies $f' \prec f$ for a non-goal valuation $f$,
  if $(f,f')$ is compatible with a policy rule in $\pi_\Phi$.
\end{theorem}

Here a pair of feature valuations $(f,f')$ is compatible with a rule $C \rightarrow E$,
if $C$ is true in $f$ and the change in values from $f$ to $f'$ is compatible with $E$.
Notice that if a state transition $(s,s')$ is compatible with $C \rightarrow E$,
the pair of feature valuations $(f(s),f(s'))$ is compatible with the rule,
but the existence of such a state transition is not a requirement for
the pair $(f,f')$ to be compatible with the policy rule;
they can be arbitrary feature valuations over $\Phi$.

\Proof{
  We must show that $\prec$ is irreflexive and well founded.
  For the first, assume that there is a sequence of feature valuations
  $f_1 \prec f_2 \prec \cdots \prec f_n$ with $n>1$ and $f_1 = f_n$.
  Without loss of generality, we may assume that the pair $(f_i,f_{i+1})$
  is compatible with a policy rule (i.e., $f_{i+1} \prec f_{i}$ is not the
  result of the transitive closure).
  In such a case, the boolean valuations $b_i$ associated with the feature
  valuations $f_i$ must form a cycle in the policy graph, and therefore
  from the termination condition, some variable must be decreased and not
  increased in the cycle, which contradicts $f_1=f_n$.

  For well-foundness, let us assume that there is an infinite chain
  $f_1 \succ f_2 \succ \cdots$ where $f_i \succ f_{i+1}$ stands for
  $f_{i+1} \prec f_i$, and let us assume, without loss of generality,
  that $(f_i,f_{i+1})$ satisfies a policy rule.
  Consider the sequence boolean valuations $\{b_i\}_{i}$ associated with
  the feature valuations $\{f_i\}_{i}$, let $b$ be a boolean valuation
  that appears infinitely often in the sequence, and let $B$ be the set
  of edges $b \rightarrow b'$ in the policy graph that are traversed
  infinitely often in the sequence.
  Then, by definition of $B$, there is an infinite subsequence $\{i_j\}_{j\geq 0}$
  of indices such that $b_{i_j}=b$ and all the edges traversed in the sequence
  $b_{i_j},\ldots,b_{i_{j+1}}$ belong to $B$, for $j\geq 0$.
  By definition of termination, there is a numerical feature $n$ that is
  decremented in some edge in $B$ but not incremented by any edge in $B$.
  Yet this is impossible, since the numerical feature $n$ is integer valued,
  non-negative, it is decremented infinitely often, and never incremented.
  The contradiction comes for supposing the existence of the infinite
  chain $f_1 \succ f_2 \succ \cdots$.
  Therefore, $\prec$ is a well founded strict partial order.
}

%% An important consequence of this theorem, exploited below, is that it suggests sets
%% of {terminating} policy rules as a \textbf{general language for expressing serializations.}
%% If the policy rules are sound and solve $\Q$, the induced serialization has width $0$, but if not,
%% the width can be greater than zero and yet bounded and small.
%%
There are simple syntactic conditions that ensure that a set of policy rules
and the graph that they define are terminating \cite{bonet:qnps}.
For example: a)~each policy rule decreases some feature and increases none,
b)~there is an ordering $n_1, \ldots, n_k$ of the numerical features such that
each policy rule decreases a feature $n_i$, while possibly increasing features $n_j$, $j > i$, and
c) a further generalization where b) holds except in some policy rules that affect boolean
features only, which cannot produce a cycle of truth valuations over the boolean features
without involving a policy rule that affects some numerical variable. If the set of policy
rules complies with the conditions a) or b), the rules are said to be \textbf{regular};
if they comply with c), the rules are said to be \textbf{weakly regular}.
In either case, these are sufficient syntactic conditions that ensure termination;
unlike the conditions captured by Definition~\ref{def:termination}, they are not necessary.

%% Since such policy rules define policy graphs that are terminating,
%% by Theorem~\ref{thm:terminating:serialization}, they define well-formed serializations.

\smallskip

\begin{example}
  The policy $\pi_\Phi$ for Boxes with rules $\prule{\GT{m}}{\DEC{m}}$
  and $\prule{\EQ{m},\GT{n}}{\DEC{n},\UNK{m}}$ and goal $\EQ{n}$ is
  \textbf{closed} and \textbf{regular}.
  It is closed since for any state $s$ where $\GT{n}$, there is a
  transition $(s,s')$ that is compatible with $\pi_\Phi$: if $\EQ{m}$,
  $s'$ results from an action that puts an empty box away, while if $\GT{m}$,
  $s'$ results from an action that puts a marble away from a box with
  a least number of marbles. % (ties broken arbitrarily).
  Theorem~\ref{thm:structurally} implies that $\pi_\Phi$ solves $\Q_B$.
\end{example}

\smallskip

If a terminating policy for the class $\Q$ defines a serialization over $\Q$,
a terminating policy that solves $\Q$ should define a serialization over $\Q$ with
$0$ width. Two conditions are needed for this though.
The first is the notion of \textbf{goal connectedness:}

\begin{definition}[Goal connected]
  \label{def:policy:goal-connected}
  A policy $\pi_\Phi$ for $\Q$ with goal separating features and its policy
  graph are said to be \textbf{goal connected}
  %% KEEP ONLY MORE GEN CONDITION
  %% when all nodes $b$ in the policy graph are connected to goal nodes, or more generally,
  when all nodes $b(f(s_0))$ associated with the initial states $s_0$ of
  instances $P$ in $\Q$ are connected only to nodes $b$ that are connected with goal nodes.
\end{definition}

Clearly, a policy $\pi_\Phi$ for $\Q$ is not closed if its policy graph
is not goal-connected, but goal-connectedness does not imply that the policy
is closed. For this, we need a condition that goes beyond the structure of
the policy graph:\footnote{The notion of soundness is similar to action
  soundness in QNPs when QNPs are used to abstract classes of problems
  \cite{bonet:ijcai2018,bonet:aaai2019}.
}

\begin{definition}[Sound policy]
  \label{def:policy:sound}
  A policy $\pi_\Phi$ over $\Q$ is \textbf{sound} if for any reachable
  non-goal state $s$ in an instance $P$ in $\Q$ where the policy rule
  $C \mapsto E$ is applicable (i.e., where $C$ holds), there is a
  transition $(s,s')$ in $P$ that is compatible with $\pi_\Phi$.
\end{definition}

Soundness and goal connectedness imply that a policy is closed,
and both are critical for establishing the conditions under which the
serialization induced by a terminating policy has zero width:

\begin{theorem}
  \label{thm:sound}
  If $\pi_\Phi$ is sound and goal-connected in $\Q$, then $\pi_\Phi$ is closed in $\Q$.
\end{theorem}

From Theorem~\ref{thm:structurally}, it follows that a terminating policy $\pi_\Phi$ that
is sound and goal-connected in $\Q$, solves $\Q$. In such a case, we say that the policy
$\pi_\Phi$ solves the class of problems $\Q$ \textbf{structurally}, as two of the conditions,
termination and goal-connectedness, can be tested on the policy graph.
Soundness, on the other hand, is the condition that ensures that only sink nodes in the policy
graph over any instance $P \in \Q$ are the goal nodes.

\Proof{
  If $\pi_\Phi$ is not closed on some instance $P$ in $\Q$,
  there is a non-goal state $s$ in $P$ that is reachable from the
  initial state $s_0$ using $\pi_\Phi$, but there is no transition
  $(s,s')$ that is compatible with $\pi_\Phi$.
  The node $b(f(s_0))$ is connected in the policy graph
  to $b(f(s))$, which from the definition of goal connectedness,
  must be connected to a goal node.
  This implies that there is an edge $b(f(s)) \rightarrow b'$
  in the policy graph, and therefore there is some policy rule
  $C \mapsto E$ that is applicable at $s$.
  Soundness of $\pi_\Phi$ then implies that there is a
  transition $(s,s')$ that is compatible with $\pi_\Phi$, a contradiction.
}

As expected, if a terminating policy $\pi_\Phi$ solves $\Q$, the width of $\Q$ under the
serialization determined by the policy (Theorem~\ref{thm:terminating:serialization}) is $0$,
provided however, that certain structural conditions hold:

\begin{theorem}
  \label{thm:structurally:w0}
  Let $\pi_\Phi$ be a policy that solves $\Q$ \textbf{structurally} and
  let $(\Phi,\prec)$ be the serialization over $\Q$ determined by $\pi_\Phi$.
  If $\pi_\Phi$ is sound and goal connected, $w_\Phi(\Q)=0$.
\end{theorem}
\Proof{
  By definition of serialized width, for any problem $P$ in $\Q$, we need to
  show that the width of any subproblem $P[s,\prec]$ in $P[\prec]$ is zero.
  For this, it is enough to show that for any such state $s$, there is a
  transition $(s,s')$ that is compatible with $\pi_\Phi$ because then, by
  definition, $f(s') \prec f(s)$ and $P[s,\prec]$ has width equal to zero.

  If $P[s_0,\prec]$ is in $P[\prec]$ for the initial state $s_0$ of $P$,
  $s_0$ is not a goal state and there is some transition $(s_0,s')$ that
  is compatible with $\pi_\Phi$ since $\pi_\Phi$ solves $P$.
  Let us now consider the subproblem $P[s',\prec]$ for $s'\neq s_0$.
  By definition, there is a subproblem $P[s,\prec]$ in $P[\prec]$ such that
  $s'$ is a non-goal state at shortest distance from $s$ with $f(s')\prec f(s)$.
  %and no goal state of $P$ is strictly closer from $s$ than $s'$.
  By definition of $\prec$ and reasoning inductively, there is a path in the
  policy graph for $\pi_\Phi$ from the boolean feature valuation $b(f(s_0))$
  to $b(f(s'))$.
  Since the features in $\Phi$ separate goals in $\Q$ and $\pi_\Phi$ is goal
  connected, there is a policy rule $C \mapsto E$ that is applicable at $s'$.
  Thus, by soundness, there must be a transition $(s',s'')$ in $P$ that is
  compatible with $\pi_\Phi$.
}

% omit

\begin{example}
  The policy for Boxes in the previous example is closed, goal connected, and sound.
  By Theorem~\ref{thm:structurally:w0}, it determines a serialization $(\Phi,\prec)$ of width zero,
  where $f(s)=[n(s),m(s)] \prec f(s')=[n(s'),m(s')]$ iff $n(s) < n(s')$ or
  $n(s)=n(s')$ and $m(s) < m(s')$.
\end{example}

\smallskip

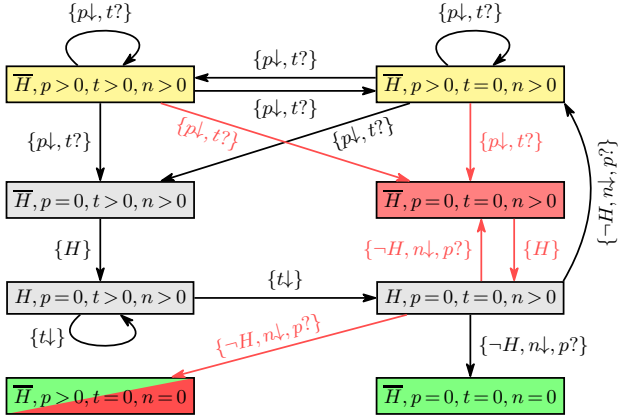
\begin{figure}
  \centering
  \resizebox{.95\columnwidth}{!}{
  \begin{tikzpicture}[thick,>={Stealth[inset=2pt,length=8pt,angle'=33,round]},font={\footnotesize},node distance=2cm,qs/.style={draw=black,fill=gray!20!white},init/.style={qs,fill=yellow!50!white},goal/.style={qs,fill=green!50!white},qa/.style={qs,fill=red!50!white}]
    %% Graph for delivery with unique target cell
    \node[init] (n0) at  (0,0)          { $\overline{H}, \GT{p}, \EQ{t}, \GT{n}$ };
    \node[qa,below = 1.25 of n0] (n2)   { $\overline{H}, \EQ{p}, \EQ{t}, \GT{n}$ };
    \node[qs,below = 1.00 of n2] (n4)   { $H, \EQ{p}, \EQ{t}, \GT{n}$ };
    \node[goal,below = 1.00 of n4] (n6) { $\overline{H}, \EQ{p}, \EQ{t}, \EQ{n}$ };

    \node[init,left = 2.90 of n0] (n1)  { $\overline{H}, \GT{p}, \GT{t}, \GT{n}$ };
    \node[qs,below = 1.25 of n1] (n3)   { $\overline{H}, \EQ{p}, \GT{t}, \GT{n}$ };
    \node[qs,below = 1.00 of n3] (n5)   { $H, \EQ{p}, \GT{t}, \GT{n}$ };
    \node[goal,below = 1.00 of n5, diagonal fill={red!70!white}{green!50!white}] (n7) { $\overline{H}, \GT{p}, \EQ{t}, \EQ{n}$ };

    \path[->] (n0) edge[out=140,in=40,looseness=4] node[above,yshift=0] { $\{ \DEC{p}, \UNK{t} \}$ } (n0);
    \path[->] (n0) edge[transform canvas={yshift=3}] node[above,yshift=0] { $\{ \DEC{p}, \UNK{t} \}$ } (n1);
    \path[->] (n0) edge[] node[sloped,xshift=35,yshift=-7] { $\{ \DEC{p}, \UNK{t} \}$ } (n3);
    \path[->] (n0) edge[red!70!white] node[right,yshift=0,red!70!white] { $\{ \DEC{p}, \UNK{t} \}$ } (n2);
    \path[->] (n2) edge[transform canvas={xshift=20},red!70!white] node[right,yshift=0,red!70!white] { $\{ H \}$ } (n4);
    \path[->] (n4) edge[] node[right,yshift=0] { $\{ \neg H, \DEC{n}, \UNK{p} \}$ } (n6);
    \path[->] (n4) edge[transform canvas={xshift=5},red!70!white] node[left,yshift=0,red!70!white] { $\{ \neg H, \DEC{n}, \UNK{p} \}$ } (n2);
    \path[->] (n4) edge[red!70!white] node[sloped,xshift=-8,yshift=7,red!70!white] { $\{ \neg H, \DEC{n}, \UNK{p} \}$ } (n7);
    \path[->] (n4) edge[transform canvas={xshift=38},bend right=30] node[sloped,xshift=0,yshift=-8] { $\{ \neg H, \DEC{n}, \UNK{p} \}$ } (n0);

    \path[->] (n1) edge[out=140,in=40,looseness=4] node[above,yshift=0] { $\{ \DEC{p}, \UNK{t} \}$ } (n1);
    \path[->] (n1) edge[transform canvas={yshift=-3}] node[below,yshift=0] { $\{ \DEC{p}, \UNK{t} \}$ } (n0);
    \path[->] (n1) edge[red!70!white] node[sloped,xshift=-35,yshift=-7,red!70!white] { $\{ \DEC{p}, \UNK{t} \}$ } (n2);
    \path[->] (n1) edge[] node[left,yshift=0] { $\{ \DEC{p}, \UNK{t} \}$ } (n3);
    \path[->] (n3) edge[] node[left,yshift=0] { $\{ H \}$ } (n5);
    \path[->] (n5) edge[] node[above,yshift=0] { $\{ \DEC{t} \}$ } (n4);
    \path[->] (n5) edge[out=220,in=320,looseness=4] node[left,xshift=-12,yshift=3] { $\{ \DEC{t} \}$ } (n5);

    %% Graph for delivery with different target cells for each package
    %% \node[init] (n0) at  (0,0)          { $\overline{H}, \GT{p}, \EQ{t}, \GT{n}$ };
    %% \node[qs,above = 1.00 of n0]   (n1) { $\overline{H}, \EQ{p}, \EQ{t}, \GT{n}$ };
    %% \node[qs,right = 2.90 of n1]   (n2) { $H, \EQ{p}, \GT{t}, \GT{n}$ };
    %% \node[qs,right = 2.90 of n0]   (n3) { $H, \EQ{p}, \EQ{t}, \GT{n}$ };
    %% \node[goal,below = 1.00 of n3] (ng) { $\overline{H}, \EQ{p}, \EQ{t}, \EQ{n}$ };
    %% \path[->] (n0) edge[] node[left,yshift=0] { $\{ \DEC{p} \}$ } (n1);
    %% \path[->] (n1) edge[] node[above,yshift=0] { $\{ H, \INC{t} \}$ } (n2);
    %% \path[->] (n2) edge[] node[right,yshift=0] { $\{ \DEC{t} \}$ } (n3);
    %% \path[->] (n3) edge[] node[above,yshift=0] { $\{ \neg H, \DEC{n}, \UNK{p} \}$ } (n0);
    %% \path[->] (n3) edge[] node[right,yshift=0] { $\{ \neg H, \DEC{n}, \UNK{p} \}$ } (ng);
    %% \path[->] (n0) edge[out=220,in=320,looseness=4] node[below,yshift=0] { $\{ \DEC{n} \}$ } (n0);
    %% \path[->] (n2) edge[out=140,in=40,looseness=4] node[above,xshift=0] { $\{ \DEC{t} \}$ } (n2);
  \end{tikzpicture}
  }
  \caption{Policy graph for Delivery for the policy defined by the rules
    $\prule{\neg H,\GT{p}}{\DEC{p},\UNK{t}}$, $\prule{\neg H, \EQ{p}}{H}$,
    $\prule{H,\GT{t}}{\DEC{t}}$, and $\prule{H,\GT{n},\EQ{t}}{\neg H, \DEC{n}, \UNK{p}}$.
    Yellow and green nodes denote initial and goal nodes respectively.
    Red nodes and edges stand for nodes and transitions in the policy graph
    that do not arise in the instances.
    The graph is terminating and goal connected, and the policy is
    closed and sound for the classes $\Q_D$ and $\Q_{D_1}$.
  }
  \label{fig:delivery}
\end{figure}

\begin{example}
  In Delivery, we use the features $\Phi=\{H,p,t,n\}$
  and the policy $\pi_\Phi$ defined by the rules
  $\prule{\neg H,\GT{p}}{\DEC{p},\UNK{t}}$, $\prule{\neg H, \EQ{p}}{H}$,
  $\prule{H,\GT{t}}{\DEC{t}}$, and $\prule{H,\GT{n},\EQ{t}}{\neg H, \DEC{n}, \UNK{p}}$.
  %$\prule{\neg H, \GT{p}}{\DEC{p}}$, $\prule{\neg H, \EQ{p}}{H, \INC{t}}$,
  %$\prule{H, \GT{t}}{\DEC{t}}$, and $\prule{H,\GT{n}, \EQ{t}}{\neg H, \DEC{n}, \UNK{p}}$.
  It is easy to check that $\pi_\Phi$ is sound for the classes $\Q_D$ and
  $\Q_{D_1}$ since for any reachable non-goal state $s$ in a problem $P$,
  there is a transition $(s,s')$ that is compatible with $\pi_\Phi$.
  On the other hand, the policy graph for $\pi_\Phi$, depicted in
  Fig.~\ref{fig:delivery}, is clearly terminating and goal connected.
  Since $\Phi$ separates goals for the classes $\Q_D$ and $\Q_{D_1}$, by
  Theorem~\ref{thm:structurally}, $\pi_\Phi$ solves both classes structurally,
  and the induced serialization $(\Phi,\prec)$ has width zero for the classes
  $\Q_D$ and $\Q_{D_1}$ by Theorem~\ref{thm:structurally:w0}.
\end{example}

% omit of whole section

\section{Sketches: A Language for Serializations}

The notion of serialization plays a key role in the connection between
general policies and serialized width, and since width is a special case of
serialized width,\footnote{$w(P)=w_\Phi(P)$ for the empty serialization
  $(\Phi,\prec)$; i.e., the one for which $f \prec f'$ is always false.
}
serializations emerge as a central concept in the study.
Indeed, serializations can be used to reduce the width of a problem, and if
this width is reduced to zero over $\Q$, a general policy for $\Q$ results.
We focus next on a specific \textbf{language} for specifying serializations
in a compact manner.
The language can be used either to \textbf{encode serializations} by hand, as a form
of domain-specific knowledge for extending the scope of width-based algorithms
such as SIW, or for \textbf{learning serializations} from traces.
Such uses, however, are beyond the scope of this paper.
%%% , very much like policies \cite{bonet:aaai2019}.
%%% These uses, however, are beyond the scope of this paper, and we thus focus
%%% on the language, semantics, and some basic results and illustrations.
%%%\cite{khardon:action,martin:generalized,fern:generalized,josh:logical,bonet:aaai2019}.

A \textbf{policy sketch} or simply \textbf{sketch} $R_\Phi$, for a class of
problems $\Q$, is a set of policy rules over the features $\Phi$ that distinguish
the goals of $\Q$. The sketch $R_\Phi$ can be a full fledged policy over $\Q$,
part of it, or just set of policy rules $C \rightarrow E$, including the empty set.
By interpreting $R_\Phi$ as a policy, we can transfer previous results that involve
policy graphs and termination. We call the rules in a sketch $R_\Phi$,
\textbf{sketch rules} because their \textbf{semantics} is different from the
semantics of policy rules.

\begin{definition}[Sketch]
  \label{def:sketch}
  A sketch for $\Q$ is a set $R_\Phi$ of sketch rules $C \rightarrow E$ over
  features $\Phi$ that \textbf{separate goals} in $\Q$.
  The sketch $R_\Phi$ is \textbf{well-formed} if the set of rules $R_\Phi$
  interpreted as a policy is \textbf{terminating}.
\end{definition}

Notice that the definition of terminating policies does not require the
policy to be closed or even to solve $\Q$.
Theorem~\ref{thm:terminating:serialization} directly yields:

\begin{theorem}
  \label{thm:sketch:serialization}
  A well-formed sketch $R_\Phi$ for $\Q$ defines a serialization $(\Phi,\prec)$ over $\Q$
  where `$\prec$' if the smallest strict partial order that satisfies $f' \prec f$ if
  the pair of feature valuations $(f,f')$ is compatible with a sketch rule in $R_\Phi$.
\end{theorem}

The distinction between policy and sketch rules is \textbf{semantical}, not
syntactical. A policy $\pi_\Phi$ defines a \textbf{filter on state transitions}:
a state transition is compatible with the policy if it is compatible with one of
its rules.
A sketch $R_\Phi$, on the other hand, is not to be used in this way: a well-formed
sketch defines a serialization, and a sketch rule $C \mapsto E$ defines
\textbf{subproblems of the serialization}: the subproblem of going from a state
$s$ where $C$ holds in $f=f(s)$ to a state $s'$ with feature valuation $f'=f(s')$
such that the pair $(f,f')$ is compatible with the sketch rule.
The key difference is that this subproblem is not solvable in a single step in general.
Theorems~\ref{thm:termination} and \ref{thm:terminating:serialization} about policy
rules that are terminating and that induce well-formed serializations are valid for
sketch rules, because in both cases, the relevant notions, like sketch graphs, are
defined structurally, without considering the state transitions that are possible in
the target class $\Q$.
Another way to see the difference between policies and sketches
is that (terminating) policies $\pi_\Phi$ play \textbf{two different roles} in our analysis:
they specify \textbf{control}, i.e., the possible actions to be done in a given instance $P$ of $\Q$,
and they define \textbf{serializations}. Sketches, on the other hand, play the latter role only.

%%% Indeed, we can define the width of a sketch rule $C \mapsto E$, although the key complexity measure is given by the
%%% width $w_\Phi(\Q)$ of the serialization $(\Phi,\prec)$ that results from the rules in the sketch and the goals of the problems in $\Q$.

Sketches provide a language for decomposing a problem into subproblems
and thus for reducing its width, which goes well beyond the language of
goal counters and variations, as the language for sketches includes the
language of general policies.

\smallskip

\begin{example}
  The serialization given by the single feature $\#g$ that counts the number
  of unachieved (top) goals is captured with the sketch that only contains
  the rule $\prule{\GT{\#g}}{\DEC{\#g}}$ when there are no other features,
  and the rule $\prule{\GT{\#g}}{\DEC{\#g},\UNK{p},\UNK{n}}$ when $p$
  and $n$ are other features.
  The rules say that it is ``good'' to decrease the goal counter independently
  of the effects on other features.
  %%% The sketch that captures the straightforward goal counter $\#g$ serialization
  %%% is given by a sketch with a single rule $\prule{\GT{\#g}}{\DEC{\#g}}$ if
  %%% there are not other features, and with the rule $\prule{\GT{\#g}}{\DEC{\#g},\UNK{p},\UNK{n}}$
  %%% if there are other boolean and numerical features $p$ and $n$, meaning
  %%% that that it's good to decrease the goal counter no matter the effects
  %%% on the other features.
  In problems such as Blocksworld, this serialization does not work (has unbounded width),
  but serializing instead with the single feature $\#m$ that counts the number
  of misplaced blocks with the sketch rule $\prule{\GT{\#m}}{\DEC{\#m}}$,
  yields a serialization of width $2$. A block is misplaced when it is on a wrong
  block or is above a mislaced block.
  %well placed when the object below it, another block or table,
  %matches the goal description, and in the case of a block, it is a well-placed block.
  The sketch thus decomposes any Blocksworld problem into subproblems
  whose goals are to put a block away or to put them at the right place.  
  The width of the first subproblem is 1 while for second is 2.
  %%% Another common serialization is via a sequence of subgoals $p_1, \ldots, p_n$,
  %%% where the idea is to be moving forward in this chain of features, while
  %%% ignoring %%% the side effects on the previous features of the sequence.
  %%% The corresponding sketch in this case is given by the rules
  %%% $\prule{\neg p_i, \ldots, \neg p_n}{p_i,\UNK{p_1}, \ldots, \UNK{p_{i-1}}}$,
  %%% $i=1, \ldots, n$.
\end{example}

\smallskip

Our last results are about the width of the serializations defined by sketches,
and the modifications in the \siw algorithm to work with sketches:

\Omit{
  The subproblems associated with serialization $(\Phi,\prec)$ for class $\Q$
  are the subproblems $P[s,\prec]$ that belong to $P[\prec]$ for any problem
  $P$ in $\Q$, each one with the goal of finding a state $s'$ reachable from $s$
  that is either a goal state or $f(s')\prec f(s)$. These subproblems may be
  hard to characterize syntactically depending on the strict partial order $\prec$.
  On the other hand, the serialization $(\Phi,\prec)$ induced by a sketch
  $R_\Phi$ can be characterized structurally.
  Indeed, the subproblem $P[s,\prec]$ associated with a state $s$ for problem
  $P$ is the problem of finding a state $s'$ reachable from $s$ such that $s'$
  is a goal state, or the pair $(s,s')$ is compatible with some sketch rule
  $r: C\mapsto E$ in $R_\Phi$.
  Therefore, sketches provide a useful method to decompose problems into
  subproblems, to bound the width of such decompositions, and to solve
  them using SIW.

  \begin{definition}
    \label{def:sketch:width}
    The width of a well-formed sketch $R_\Phi$ over a class $\Q$ of problems
    is the width of the serialization $(\Phi,\prec)$ induced by $R_\Phi$ on $\Q$.
    In particular, the width of subproblem $P[s,\prec]$ is the width of the
    problem of to find starting at $s$ some state $s'$ reachable from $s$
    such that $s'$ is a goal state or $(s,s')$ is compatible with some
    sketch rule in $R_\Phi$.
  \end{definition}
} % omit

\begin{definition}[Sketch width]
  \label{def:sketch:width}
  Let $R_\Phi$ be a well-formed sketch for a class of problems $\Q$ such that $\Phi$
  separates the goals, and let $s$ be a reachable state in some instance $P$ of $\Q$.
  The width of the sketch $R_\Phi$ \textbf{at state $s$ of problem $P$,} $w_{R}(P[s])$,
  is the width of the subproblem $P[s]$ that is like $P$ but with initial
  state $s$ and goal states $s'$ such that $s'$ is a goal state of $P$, or the pair
  $(f(s),f(s'))$ is compatible with a sketch rule $C \mapsto E$.
  The \textbf{width of the sketch} $R_\Phi$, $w_{R}(\Q)$, is the maximum
  width $w_{R}(P[s])$ for any reachable state $s$ in any problem $P$ in $\Q$.
\end{definition}

\begin{theorem}
  \label{thm:sketch:width}
  Let $R_\Phi$ be a well-formed sketch for a class $\Q$ of problems, and let $(\Phi,\prec)$
  be the serialization determined by $R_\Phi$ from Theorem~\ref{thm:sketch:serialization}.
  The width $w_\Phi(\Q)$ of the serialization is bounded by the width $w_{R}(\Q)$
  of the sketch.
\end{theorem}
\Proof{
  By definition, $w_\Phi(\Q)\leq k$ if the width $w(P[s,\prec])$ of the subproblems
  $P[s,\prec]$ in $P[\prec]$ is bounded by $k$, for any $P$ in $\Q$.
  The goals of subproblem $P[s,\prec]$ are the states $s'$ that are a goal of $P$,
  or $f(s')\prec f(s)$.
  In particular, if the pair $(f(s),f(s'))$ is compatible with a sketch rule, % $C\mapsto E$,
  then $f(s')\prec f(s)$, but the converse does not hold in general.
  That is, for any subproblem $P[s,\prec]$ in $P[\prec]$, the goals of the subproblem $P[s]$ are
  also goals of $P[s,\prec]$.
  Hence, $w_\Phi(P[s,\prec]) \leq w_{R}(P[s])$ and $w_\Phi(\Q)\leq w_{R}(\Q)$.
}

If a well-formed sketch $R_\Phi$ has bounded width for a class of problems $\Q$,
then the problems in $\Q$ can be solved in polynomial time by the algorithm \siwR
that is like \siw, with the difference that the precedence test $f \prec f'$ among
pairs of feature valuations $f$ and $f'$ is replaced by the test of whether
the feature valuation pair $(f,f')$ is compatible with a rule in $R_\Phi$. In other words,
\siwR start at the state $s:=s_0$, where $s_0$ is the initial state of $P$,
and then performs an IW search from $s$ to find a state $s'$ that is a goal state
or such the pair $(f(s),f(s'))$ is compatible with a sketch rule in $R_\Phi$.
Then if  $s'$ is not a goal state, $s$ is set to $s'$, $s := s'$, and the loop
repeats until a goal state is reached. The precedence test in $R_\Phi$ can be done in constant
time unlike the general test $f' \prec f$ in \siw.\footnote{For testing $f'\prec f$, one 
  needs to check if there is a sequence $\{f_i\}_{i=0}^n$ of feature valuations such that $f_0=f$,
  $f_n=f'$, and each pair $(f_i,f_{i+1})$ is compatible with a sketch rule, $i=0,1,\ldots,n-1$.
  However, this test can be done in constant time too, provided that the binary relation `$\prec$'
  for each instance $P$ is precompiled in a boolean hash table with $N^k$ rows and $N^k$ columns
  where $N$ is the number of atoms in $P$, $k$ is the number of features, and
  $N^k$ is the number of feature valuations in $P$.
  Unlike the procedure \siwR, this precompilation is not practical in general.
  The efficiency of \siwR comes at a price: by testing ``progress'' with the sketch rules directly
  and not with the serializations that results from such rules, \siwR is not using the serialization
  at full, as it ignores the transitive closure of the precedence relations.
}
The runtime properties of \siwR are thus similar to those of \siw, as captured in Theorem~\ref{thm:serialization:siw},
with precedence tests that can be done in constant time:

\begin{theorem}
  \label{thm:sketch:siw}
  Let $R_\Phi$ be a well-formed sketch for a class $\Q$ of problems.
  If the sketch width $w_{R}(\Q)$ is bounded by $k$, \siwR solves any problem $P$
  in $\Q$ in $O(bN^{|\Phi|+2k-1})$ time and $O(bN^k + N^{|\Phi|+k})$ space,
  where $b$ and $N$ bound the branching factor and number of atoms in $P$ respectively.
\end{theorem}
\Proof{
  Like the proof of Theorem~\ref{thm:serialization:siw} but
  with the test $f \prec f'$ in \siw replaced by checking if
  the pair of feature valuations $(f,f')$ is compatible with a rule in $R_\Phi$.
}

% omit

\begin{example}
  Different and interesting sketches are given in Table~\ref{table:sketches} for the
  two classes of problems for Delivery: the class $\Q_D$ of problems with an arbitrary
  number of packages and the class $\Q_{D_1}$ of problems with a single package.
  In the table, the entries in the columns $\Q_{D_1}$ and $\Q_{D}$ upper bound the width
  of the different sketches in the table for the two classes of Delivery problems.
  The entries $unb$ and `---' stand respectively for unbounded width and ill-defined
  (non-terminating) sketch.
  The features used are: (boolean) $H$ for holding a package, $p$ is distance to nearest
  package (zero if holding a package or no package to be delivered remains, $t$ is
  distance to current target cell (zero if holding nothing), and $n$ is number of
  packages still to be delivered.

  \begin{table}
    \begin{center}
      %\resizebox{\columnwidth}{!}{
        \begin{tabular}{@{}lcc@{}}
          \toprule
          Policy sketch                                                                   & $\Q_{D_1}$ & $\Q_D$ \\
          \midrule
          $\sigma_0=\{\}$                                                                 &         2 &  $unb$ \\
          $\sigma_1=\{\prule{H}{\neg H,\UNK{p},\UNK{t}}\}$                                &         2 &  $unb$ \\
          $\sigma_2=\{\prule{\neg H}{H,\UNK{p},\UNK{t}}\}$                                &         1 &  $unb$ \\
          $\sigma_3=\sigma_1 \cup \sigma_2$                                               &       --- &    --- \\
          $\sigma_4=\{\prule{\GT{n}}{\DEC{n},\UNK{H},\UNK{p},\UNK{t}}\}$                  &         2 &      2 \\
          %%$\sigma_4=\sigma_1 \cup \{\prule{H,\GT{n}}{\neg H,\UNK{t},\UNK{p},\UNK{n}}\}$ &       --- &    --- \\
          $\sigma_5=\sigma_2 \cup \sigma_4$                                               &         1 &      1 \\
          $\sigma_6=\{\prule{\neg H, \GT{p}}{\DEC{p}, \UNK{t}}\}$                         &         2 &  $unb$ \\
          $\sigma_7=\{\prule{H, \GT{t}}{\DEC{t},\UNK{p}}\}$                               &         2 &  $unb$ \\
          $\sigma_8= \sigma_2 \cup \sigma_4 \cup \sigma_6 \cup \sigma_7$                  &         0 &      0 \\
          \bottomrule
        \end{tabular}
      %}
    \end{center}
    \caption{Upper bounds on the width of different sketches for the classes $\Q_{D_1}$ and $\Q_{D}$ of Delivery problems.
      The entries $unb$ and `---' mean, respectively, unbounded width and ill-defined sketch.
      For sketches of bounded width, \siwR solves any instance in the class in polynomial time.
    }
    \label{table:sketches}
  \end{table}

  We briefly explain the entries in the table without providing formal proofs
  (such proofs can be obtained with Theorem~\ref{thm:projection:class}).
  $\sigma_0$ is the empty sketch whose width is the same as
  the plain width, 2 for $D_1$ and unbounded for $D$, as no problem $P$
  is decomposed into subproblems.
  The rule $\prule{H}{\neg H,\UNK{p},\UNK{t}}$ in $\sigma_1$ does not
  help in initial states that do not satisfy $H$, and hurts a bit
  in states that do satisfy $H$. Indeed, in the latter states $s$,
  there is a state $s'$ at one step ahead from $s$ with $f(s') \prec f(s)$
  (obtained from $s$ by dropping the package) that gets chosen by
  any algorithm like \siwR.
  However, in the resulting subproblem
  with initial state $s'$ there is no state $s''$ for which $f(s'') \prec f(s')$
  holds as there is no sketch rule $C \mapsto E$ where $C$ is true in $s'$.
  As a result, the goal of the subproblem rooted as $s'$ is the true goal
  of the problem, which may actually involve going back from $s'$ to $s$
  and from $s$ to the goal. This is indeed what \siwR\ does given %the sketch
  $\sigma_1$. Since this subproblem has width $2$, the serialized
  width of $D_1$ remains $2$.
  For $\sigma_2$, the rule $\prule{\neg H}{H,\UNK{p},\UNK{t}}$ says that
  a state $s$ where $\neg H$ holds can be ``improved'' by finding a state
  $s'$ where $H$ holds, while possibly affecting $p$, $t$, or both.
  Hence, any problem $P$ in $D_1$ is split in two subproblems: achieve $H$
  first and then the goal, reducing the sketch width of $D_1$ to 1 but not
  the sketch width of $D$.
  The sketch $\sigma_3$ is not well-formed as it is not terminating: indeed
  the resulting ordering is not a strict partial order: a state $s$ where
  the agent is at the same location as a package but does not hold it is
  ``improved'' into a state $s'$ by picking the package, which is improved
  again by dropping the package without affecting any numerical feature.
  For $\sigma_4$, that subgoals on the top goal captured by the feature $n$, that counts
  the number of undelivered packages, the sketch width of $D$ reduces to 2 but
  that of $D_1$ remains unchanged. %does not affect the width of $D_1$;
  % i.e., a problem in $D$ is split into $n$ subproblems each of ``type'' $D_1$.
  %
  The sketch $\sigma_5$ combines the rules in $\sigma_2$ and $\sigma_4$,
  and as a result, decomposes $D$ into width 2 subproblems, the first of which,
  that corresponds to $D_1$, is further decomposed into two subproblems.
  The sketch width of both $D$ and $D_1$ becomes then 1: the first subproblem is
  to collect the nearest package, the second one to drop it at the target cell,
  and the same sequence of subproblems gets repeated.
  The sketch $\sigma_6$ renders the subproblem of getting close to the nearest
  package with width $0$, but the remaining subproblem in $D_1$ that involves picking up
  the package and delivering it at the target cell, still has width 2.
  The sketch $\sigma_7$ does not decompose the problem when $H$ is initially false.
  Finally, the combination in $\sigma_8$ %of four of these sketches
  yields a full policy, and thus a serialization of width 0
  % (cf.\ Theorem~\ref{thm:structurally:w0}) %% not sure if the theorem applies
  where each subproblem is solved in a single step.
  \Omit{
    Finally, the sketch $\sigma_5$ decomposes $D_1$ into $1+p$ subproblems
    of width 1 ($p$ subproblems that reduce $p$ down to zero, and an additional
    subproblem of delivering the single package once the agent is at package's
    location) but it cannot decompose $D$ into subproblems of bounded width.
    On the other hand, $\sigma_6$ does not decompose either $D$ or $D_1$
    because for the initial state $s_0$ where $\neg H$ holds, the goal of
    the subproblem $P[s_0,\prec]$ is the goal of $P$.
    Yet, if we combine the rules in these last two sketches with those in $\sigma_1$
    and $\sigma_2$ we get a sketch of width $0$ by Theorem~\ref{thm:structurally:w0},
    from which a complete policy can be constructed using Theorem~\ref{thm:sketch:w0}.
  } % omit
\end{example}

\section{Conclusions}

We have established a number of connections between the notions of width, as
developed and used in classical planning, and the notion of generalized plans,
which are summarized in Table~\ref{table:results}.
The results suggest a deep connection between these two notions and that bounded width for infinite
collections of problems $\Q$ is often the result of simple general policies
that solve $\Q$ optimally in terms of features that are partially represented in the problem encodings.
When this is not the case, we have shed light on the representations that deliver
such properties, and hence, polynomial-time searches.
We have also formalized and generalized the notion of serialized width by appealing
to an explicit and abstract notion of serializations, and established connections
between generalized policies and
serialized width by borrowing notions from QNPs. Moreover, from this connection,
we introduced policy sketches which make use of the language of policy rules but
with a different semantics for providing a convenient and powerful language for specifying
subproblems and serializations that can be exploited by algorithms such as SIW.
The language can be used for encoding
domain-specific knowledge by hand, or as a target language for learning domain
serializations automatically from traces. These are interesting challenges
that we would like to address in the future.

\begin{table}
  \centering
  \resizebox{\columnwidth}{!}{
    \begin{tabular}{@{}cp{\columnwidth}@{}}
      \toprule
      Thm & Notes \\
      \midrule
      \ref{thm:width:nir} & Performance and guarantees of \iw{k}. \\
      \ref{thm:markovian:width} & Optimal and Markovian policies bound width if features encoded. \\
      %\ref{thm:markovian:width} & Optimal, Markov policies bound width if features encoded. \\
      \ref{thm:markovian:solutions} & Markovian policies guarantee optimal solutions with \iwf{\Phi}. \\
      \ref{thm:width} & Bounded width also when tuples capture the features in all optimal trajectories. \\
      %\ref{thm:theta} & Relation between admissible chains and the policies defined by them. \\
      \ref{thm:projection:class} & Bounded width when tuples capture features in a projection. \\
      \ref{thm:serialization:siw} & Performance and guarantees of SIW$_\Phi$. \\
      \ref{thm:termination} & Conditions for termination of policies. \\
      \ref{thm:structurally} & Closed and terminating policies define structural solutions. \\
      \ref{thm:terminating:serialization} & Terminating policies define serializations. \\
      %\ref{thm:sound} & Sound and goal-connected policies are closed. \\
      \ref{thm:structurally:w0} & Structural solutions that are sound and closed define serializations of zero width. \\
      \ref{thm:sketch:serialization} & Well-formed sketches define serializations. \\
      \ref{thm:sketch:width} & Sketch width bounds width of induced serialization. \\
      \ref{thm:sketch:siw} & Performance and guarantees of \siwR given sketches. \\
      %\ref{thm:sketch:w0} & Sketches of zero width define policies. \\
      \bottomrule
    \end{tabular}
  }
  \caption{Summary of main formal results.}
  \label{table:results}
\end{table}

\subsection*{Acknowledgments}

The research is partially funded by an ERC Advanced Grant (No 885107), by grant TIN-2015-67959-P from MINECO, Spain,
and by the Knut and Alice Wallenberg (KAW) Foundation through the WASP program.
% Wallenberg AI, Autonomous Systems and Software Program (WASP).
H.\ Geffner is also a Wallenberg Guest Professor at Link\"oping University, Sweden.
%H.\ Geffner thanks Ferran Alet and the participants of the MIT Embodied Intelligence Seminar
%for the invitation to talk and for questions regarding the notion of width that
%gave us the final push to finish this work. The research is partially funded by
%an ERC Advanced Grant (N.\ 885107), by grant TIN-2015-67959-P from MINECO, Spain,

\bibliography{control}

\end{document}